\documentclass[10pt,twocolumn,letterpaper]{article}

\usepackage{cvpr}      
\usepackage[accsupp]{axessibility} 

\usepackage{enumitem}
\usepackage{algorithm}
\usepackage{algorithmic}
\usepackage{graphicx}
\usepackage{booktabs}
\usepackage{multirow} 
\usepackage{amsfonts}
\newtheorem{theorem}{Theorem}

\newtheorem{definition}{Definition}

\usepackage{wrapfig}
\usepackage{bm}
\usepackage{comment}

\makeatletter
\newcommand*{\addFileDependency}[1]{
  \typeout{(#1)}
  \@addtofilelist{#1}
  \IfFileExists{#1}{}{\typeout{No file #1.}}
}
\makeatother

\definecolor{cvprblue}{rgb}{0.21,0.49,0.74}
\usepackage[pagebackref,breaklinks,colorlinks,allcolors=cvprblue]{hyperref}

\def\paperID{14742} 
\def\confName{CVPR}
\def\confYear{2025}

\title{Balancing Two Classifiers via A Simplex ETF Structure for Model Calibration}

\author{Jiani Ni$^{1}$\hspace{20pt}  He Zhao$^2$ \hspace{20pt} Jintong Gao$^1$\hspace{20pt} Dandan Guo$^{*1}$ \hspace{20pt} Hongyuan Zha$^3$\\ 
{$^1 $} School of Artificial Intelligence, Jilin University \hspace{20pt} {$^2 $} CSIRO's Data61 \\ 
{$^3 $} The Chinese University of Hong Kong, Shenzhen 
\\
{\tt\small \{nijn23,gaojt20\}@mails.jlu.edu.cn\hspace{20pt} c \hspace{20pt} he.zhao@ieee.org} 
}

\begin{document}
\maketitle
\begin{abstract}
In recent years, deep neural networks (DNNs) have demonstrated state-of-the-art performance across various domains. 
However, despite their success, they often face calibration issues, particularly in safety-critical applications such as autonomous driving and healthcare, where unreliable predictions can have serious consequences. Recent research has started to improve model calibration from the view of the classifier. However, the exploration of designing the classifier to solve the model calibration problem is insufficient.  
Let alone most of the existing methods ignore the calibration errors arising from underconfidence. In this work, we propose a novel method by \textbf{Bal}ancing learnable and ETF classifiers to solve the overconfidence or underconfidence problem for model \textbf{CAL}ibration named \textbf{BalCAL}. 
By introducing a confidence-tunable module and a dynamic adjustment method, we ensure better alignment between model confidence and its true accuracy. Extensive experimental validation shows that ours significantly improves model calibration performance while maintaining high predictive accuracy, outperforming existing techniques. This provides a novel solution to the calibration challenges commonly encountered in deep learning. Our code is available at \href{https://github.com/JianiNi/BalCAL}{BalCAL}.


\end{abstract}    
\section{Introduction}
\label{sec:intro}

\begin{figure*}
  \centering
  \begin{subfigure}{0.31\linewidth}
    \includegraphics[width=1.0\linewidth]{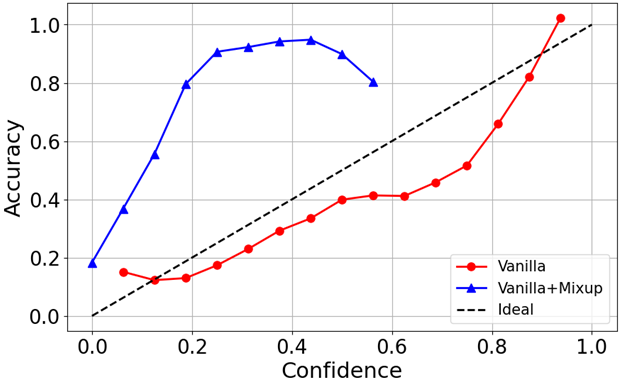}
    \caption{}
    \label{fig:motivation-a}
  \end{subfigure}
  \hfill
  \begin{subfigure}{0.36\linewidth}
    \includegraphics[width=1.0\linewidth]{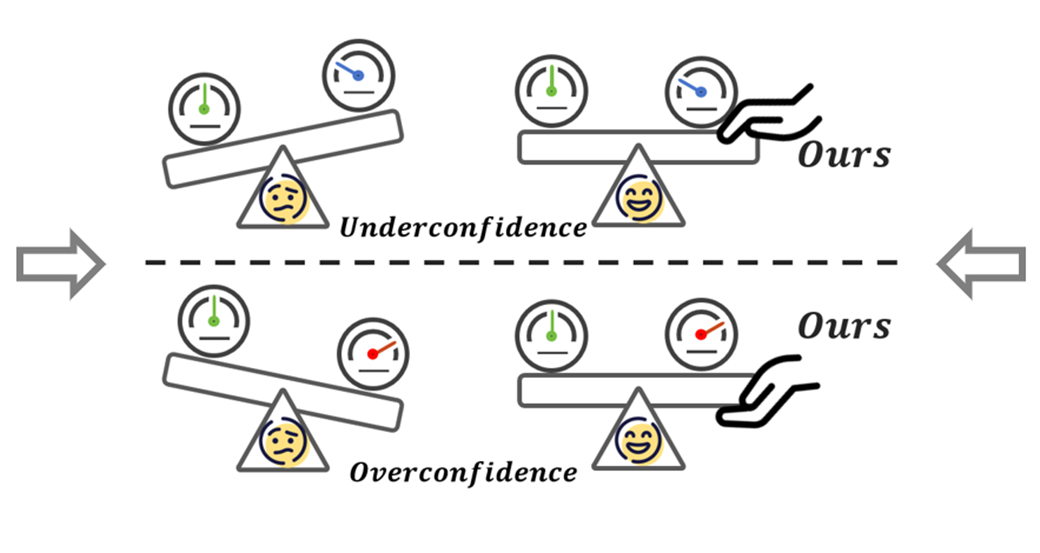}
    \caption{}
    \label{fig:motivation-b}
  \end{subfigure}
  \hfill
  \begin{subfigure}{0.31\linewidth}
    \includegraphics[width=1.0\linewidth]{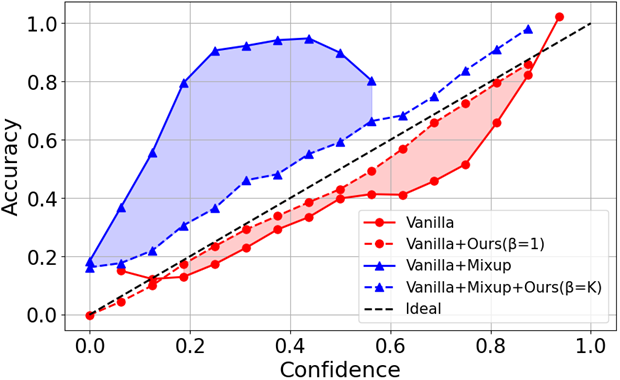}
    \caption{}
  \end{subfigure}
  \label{fig:motivation-c}
  \caption{\textbf{The motivation of BalCAL.} Calibration error arises from the discrepancy between model confidence and actual accuracy, often manifesting as overconfidence\textbf{ (a)}. After incorporating Mixup, underconfidence may also occur\textbf{ (a)}. We propose a method to dynamically adjust confidence, addressing both issues simultaneously \textbf{(b)}. Calibration performance is shown in \textbf{(c)}. }
  \label{fig:motivation}
\end{figure*}

Recently, deep neural networks (DNNs) have achieved remarkable success across various domains, including image classification \citep{he2016deep,zagoruyko2016wide,dosovitskiy2020image}, object detection \citep{he2017mask,redmon2018yolov3}, and natural language processing \citep{vaswani2017attention,devlin2018bert}. Despite their impressive performance, DNNs often exhibit calibration error, characterized by overconfidence, where the predicted probabilities exceed the accuracy of their predictions \citep{guo2017calibration}. 
This calibration error poses serious challenges in real-world risk-sensitive applications, including autonomous driving \citep{chen2017end,gupta2021deep}, healthcare diagnostics \citep{shen2019deep,ahsan2022machine} and predicting credit risk \citep{clements2020sequential} where precise confidence estimates are crucial. 

Consequently, to ensure that model predictions align more closely with their predicted accuracy, improving the confidence calibration of DNNs has become a pivotal area of research. Calibration approaches in DNNs are generally categorized into post-hoc and training-time methods. Post-hoc methods, such as Temperature Scaling (TS) \citep{guo2017calibration}, rescale output probabilities after training without modifying internal features. In contrast, training-time methods like label smoothing (LS) \citep{szegedy2016rethinking}, focal loss (FL) \citep{lin2017focal}, and Mixup \citep{zhang2018mixup} integrate calibration into the training process. Here, we focus on training-time methods.

Recent training-time methods related to classifiers have provided valuable insights. Such as the approach by \cite{jordahn2024decoupling}, aims to improve calibration by decoupling feature extraction from classification. Although this separation enhances calibration, it may reduce model flexibility and limit generalization to out-of-distribution (OOD) data, highlighting a critical trade-off between calibration and adaptability.
Additionally, \cite{wang2024calibration} found that weight decay regularization can negatively affect the model calibratability, limiting calibration improvements achievable through post-hoc methods. They proposed the weak classifier hypothesis, suggesting that a weakly trained classification head aids the representation module in generating more calibratable features. {Despite their effectiveness, it is questionable whether the exploration of designing the classifier to solve the model calibration problem is sufficient. Besides, most calibration methods primarily address overconfidence and a few researchers address calibration errors stemming from underconfidence in model predictions \citep{wang2023pitfall}. }



In this work, {we propose a novel method by fusing two classifiers, denoted as \textbf{\textit{BalCAL}}, which \textbf{\textit{Bal}}ances two classifiers to solve the overconfidence or underconfidence problem for model \textbf{\textit{CAL}}ibration. 
In addition to the learnable standard classifier, we here consider another fixed Simplex Equiangular Tight Frame (ETF) classifier from the Neural Collapse (NC) phenomenon \citep{papyan2020prevalence}. NC has emerged as a significant phenomenon in deep learning, where as training progresses the features of the final layer and the classifiers tend to converge into highly structured and symmetric patterns and classifier weights converge to a Simplex ETF structure. Recent studies \citep{yang2022inducing, peifeng2023feature} have demonstrated that Neural Collapse-inspired methods, including fixed Simplex ETF classifiers, can enhance model generalization and tackle challenges such as imbalanced learning. However, their potential for enhancing model calibration has yet to be fully explored. }

{In this work, we find that adjusting the scaling factor of the fixed ETF classifier can effectively modulate the output confidence without affecting the accuracy. That is to say, we can view the fixed ETF classifier with the tunable scaling factor as the auxiliary classifier to compensate for the predictions,  produced by the commonly used model including a feature extractor and a standard classifier. 
Specifically, during the training process, we additionally introduce a confidence-tunable module after the feature extractor, which consists of an adapter followed by the fixed Simplex ETF classifier. Considering the learnable classifier in DNNs is usually susceptible to calibration error, we design a dynamic adjustment mechanism based on the relationship between model accuracy and confidence evaluated by the training dataset to balance the standard classifier and ETF classifier. By benefiting from the combination of two classifiers, we can correct not only overconfidence but also underconfidence.
Besides, ours can be easily integrated into most previous training-time methods, offering flexibility in deployment. Our key contributions are as follows:
} 



\begin{itemize}
\item We provide a novel view that ETF classifier can be used to solve the model calibration by tuning its scaling factor.

\item  We introduce a confidence-tunable module built on the ETF classifier and design a dynamic adjustment mechanism to balance the two classifiers, denoted as \textbf{\textit{BalCAL}}, solving the overconfidence or underconfidence problem.

\item Extensive experiments demonstrate that ours achieves superior performance while achieving robustness under corruption in various settings.
\end{itemize}

\section{Related work}
\label{sec:related}

\paragraph{Confidence calibration.} 
Calibration aims to adjust predicted confidence to reflect true accuracy \citep{guo2017calibration}, primarily addressing the prevalent issue of overconfidence in deep learning models, as well as the less common but important issue of underconfidence. There are two main strategies: post-hoc and training-time methods. Temperature Scaling (TS) \citep{guo2017calibration} is a commonly employed post-hoc calibration method that adjusts the temperature of the softmax layer after training to refine the model confidence.  While TS is an effective approach, \cite{ovadia2019can} demonstrated its limitations under distribution shifts. Training-time approaches consider calibration directly during the learning process, where label smoothing (LS) \citep{szegedy2016rethinking}, focal loss (FL) \citep{lin2017focal}, and Mixup \citep{zhang2018mixup} were initially developed to improve generalization while simultaneously enhancing calibration. Some training-time methods \citep{tao2023dual,tao2023calibrating,cheng2022calibrating,ghosh2022adafocal,liu2022devil,muller2019when,pinto2022using,thulasidasan2019mixup} apply regularization to output probabilities. In this work, we primarily focus on training-time methods from the view of the classifier.


A related work is proposed by \cite{jordahn2024decoupling}, which decouples feature extraction from classification to improve calibration, including Two-Stage Training (TST) and Variational Two-Stage Training (V-TST). However, this method is limited in out-of-distribution samples, potentially causing suboptimal performance on tasks requiring broader generalization. 
Recently, \cite{wang2024calibration} observed that some training-time calibration methods may compromise the model calibratability, i.e., the ability to be further improved through post-hoc calibration. They introduced a weak classifier hypothesis and proposed Progressive Layer-Peeled (PLP) training by incrementally freezing the parameters of the upper layers to improve calibratability, thereby enhancing the performance of the model when subjected to post-hoc calibration. 
The key difference between ours and the above methods is that we introduce a fixed ETF classifier to construct a confidence-tunable module, adaptively balancing the confidence level of the learnable classifier. It also ensures robust calibration and generalization across varying data distributions.

Most training-time methods focus on mitigating overconfidence but are less effective at addressing underconfidence. For instance, Mixup improves model calibration and accuracy but is sensitive to interpolation levels, and higher levels can lead to underconfidence. 
The method proposed by \citep{wang2023pitfall} can alleviate this issue but is specific to Mixup and not generalizable. In contrast, our method effectively handles overconfidence and underconfidence, offering a general solution for model calibration. Additionally, it can be easily integrated with most of the existing calibration methods to enhance calibration performance without sacrificing accuracy, providing a flexible deployment.

\paragraph{Neural collapse.} Neural collapse is a phenomenon observed during the final stage of training on balanced datasets. In this stage, features from the last layer collapse into class means, and classifier weights converge to a Simplex Equiangular Tight Frame (ETF) structure \citep{papyan2020prevalence}. This phenomenon has sparked extensive theoretical exploration \citep{liu2023generalizing,xu2023quantifying,yang2023are,tirer2022extended,kothapalli2023neural}. More recent works have leveraged Neural Collapse-inspired methods to address issues with imbalanced training \citep{gao2024distribution, yang2022inducing,xie2023neural,thrampoulidis2022imbalance}, as well as in incremental learning \citep{yang2023neural} and transfer learning \citep{li2022understanding}. 


To our knowledge, this is the first study incorporating the ETF classifier in the NC phenomenon into the model calibration problem. By recognizing that the ETF's scaling factor directly influences predicted confidence levels, we propose a confidence-tunable module by introducing the fixed ETF classifier. This module modifies the scaling factor of the Simplex ETF, enabling self-regulation based on the model's confidence outputs. This approach enhances the model's calibration performance by correcting overconfidence or underconfidence in the model's outputs.

\section{Background}
\label{sec:back}
\begin{figure*}
    \centering
    \includegraphics[width=0.95\linewidth]{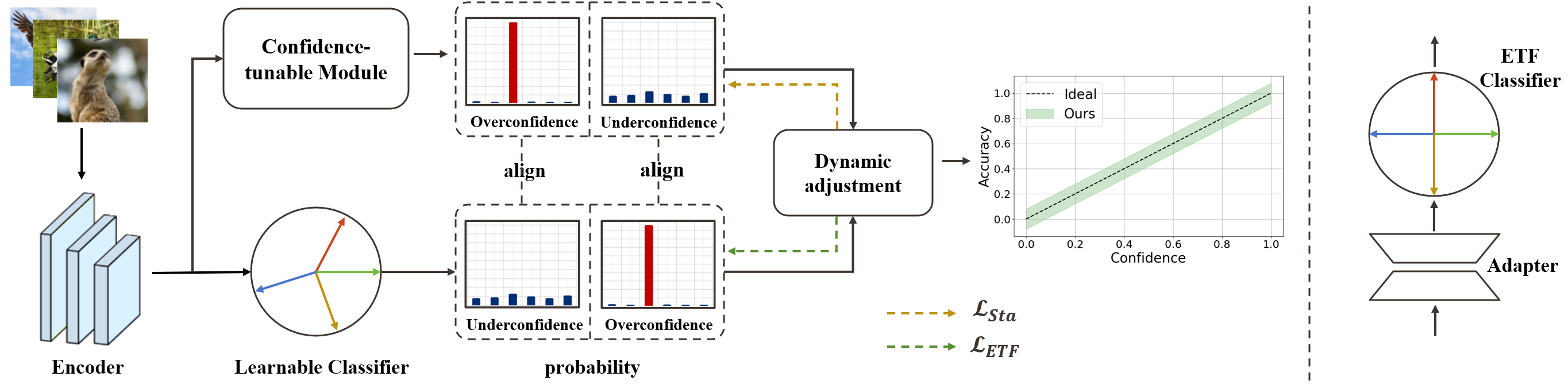}
    \caption{\textbf{An illustration of the BalCAL.} (a) The \textbf{left} side is an overview of ours, where the image is input into a shared encoder. The encoded feature is fed into the learnable classifier and the confidence-tunable module, respectively, where the latter aims to adjust its confidence to complement that of the former. A dynamic adjustment mechanism is proposed to balance the confidence between the two components. (b) The \textbf{right} side depicts the confidence-tunable module,  consisting of an adapter and a fixed ETF classifier, which works in tandem with the learnable classifier to refine the overall model’s confidence.}
    \label{fig:overall}
    \vspace{-0.3cm}
\end{figure*}
\subsection{Network calibration}
Consider a dataset \( D = \{(x_i, y_i)\}_{i=1}^{N} \), where \( x_i \in \mathcal{X} \) is input sample, \( y_i \in \mathcal{Y} \) is the corresponding label consisting of \( K \) classes,  $N$ denotes the total number of samples and  \( n_k \) is the number of samples in the \( k \)-th class with \( N = \sum_{k=1}^{K} n_k \). We model the deep learning system as consisting of two components: a feature extractor \( f: \mathcal{X} \to \mathbb{R}^d \) with parameters \( \bm{\theta} \), and a linear classifier \( \mathbf{W} = \{\bm{w}_k\}_{k=1}^{K} \in \mathbb{R}^{d \times K} \). The feature extractor generates a \( d \)-dimensional feature vector \( \bm{z}_i = f(x_i; \bm{\theta}) \in \mathbb{R}^d \), which is also referred to as the last-layer feature or activation. These features are passed to the linear classifier, which outputs the logits vector \( \bm{l}_i = \bm{z}_i \mathbf{W} \in \mathbb{R}^K \). Now the probability \( p_{i,k} \) that sample \( x_i \) belongs to class \( k \) is computed using the softmax function, denoted as $p_{i,k} = \frac{\exp(l_{i,k})}{\sum_{j=1}^{K} \exp(l_{i,j})}$.  
The predicted label and its associated confidence are defined as
\(\hat{y}_i = \arg\max_k p_{i,k}\) and \(\hat{p}_i = \max_k p_{i,k}\).  

The parameters \( \{ \bm{\theta}, \mathbf{W} \} \) are usually trained on \( D \) using standard cross-entropy loss, which tends to be overconfident \cite{guo2017calibration}. Therefore, it results in a discrepancy between predicted confidence and actual accuracy, referred to as calibration error. 
Additionally, \cite{thulasidasan2019mixup} observes that intensive interpolation in mixup training can lead to underconfidence, also leading to its calibration error. 
Calibration methods aim to align predicted confidence with actual accuracy, ensuring that \(\hat{p}_i = P(\hat{y}_i = y_i \mid \hat{p}_i)\), where \( \hat{p}_i\) reflects the true probability of a correct prediction.


\label{Network calibration}

\subsection{Neural collapse}
\label{sec:NC}
The phenomenon of neural collapse, described by \cite{papyan2020prevalence}, occurs at the final stage of training when both the class means and the linear classifier vectors converge to the vertices of a Simplex ETF, as defined below.
\vspace{-0.2cm}
\begin{definition}
\label{Simplex ETF}
\textbf{(The general Simplex ETF)}: 
{Considering the $K$-classification problem, a general} Simplex ETF is defined as a set of vectors in \( \mathbb{R}^K \) given by the columns of the matrix \( \mathbf{M}   \in \mathbb{R}^{d \times K}\):



\begin{equation}\label{ETF_classifier}
\mathbf{M} = \beta \sqrt{\frac{K}{K-1}} \mathbf{U}  ( \mathbf{I} - \frac{1}{K} \mathbf{1}_K \mathbf{1}_K^\mathrm{T} ),
\end{equation}
where $\beta \in \mathbb{R}^+$ is a scaling factor, $\mathbf{U} \in \mathbb{R}^{d \times K}$ is a partial orthogonal matrix (i.e., \( \mathbf{U}^T \mathbf{U} = I \)) with $d \geq K$, $\mathbf{I} \in \mathbb{R}^{K \times K}$ is the identity matrix, and $\mathbf{1}_K \in \mathbb{R}^K$ is an all-ones vector. Each vector in a Simplex ETF shares an identical $L2$ norm and maintains the same angle between any pair of vectors.
\end{definition}
\vspace{-0.5cm}
\paragraph{Statistics}
For a given classification task, the class mean \( \bm{z}_k \) is defined as \( \bm{z}_k = \frac{1}{n_k} \sum_{i=1}^{n_k} z_{i,k} \), where \( z_{i,k} \) denotes the feature vector of the \( i \)-th sample from the \( k \)-th class. The global mean \( \bm{z}_G \) is given by \( \bm{z}_G = \frac{1}{K} \sum_{k=1}^{K} \bm{z}_k \). The within-class covariance matrix \( \Sigma_\mathbf{W} \) quantifies the variance of feature vectors within each class, measured as the dispersion around their mean. It is expressed as \(\Sigma_\mathbf{W} = \frac{1}{K} \sum_{k=1}^{K} \sum_{i=1}^{n_k} \frac{1}{n_k} (z_{i,k} - \bm{z}_k)(z_{i,k} - \bm{z}_k)^\top\).
The class mean is first centered by subtracting the global mean and then normalized by its \(L2\) norm, resulting in \( \bm{\tilde{z}}_k = \frac{\bm{z}_k - \bm{z}_G}{\|\bm{z}_k - \bm{z}_G\|_2} \). The matrix \( \mathbf{\tilde{M}} = [\bm{\tilde{z}}_k] \in \mathbb{R}^{d \times K} \) is formed by stacking the normalized class means as columns. The symbol \( \rightarrow \) indicates the convergence of the model as training progresses. During the final phase of training, there are four interdependent characteristics associated with the following conditions:

\begin{itemize}[leftmargin=*]
    \item \textbf{(NC1) Variability Collapse: }The within-class variability diminishes, with \( \Sigma_\mathbf{W} \to 0 \).

    \item \textbf{(NC2) Convergence to Simplex ETF:} Class mean vectors converge to a Simplex ETF, with uniform \(L2\) norms and pairwise angles, so that \( \|\bm{z}_k - \bm{z}_G\| - \|\bm{z}_{k'} - \bm{z}_G\| \to 0 \) and \( \langle \bm{\tilde{z}}_k,  \bm{\tilde{z}}_{k'} \rangle \to -\frac{1}{K-1} \) for \( k \neq k' \).

    \item \textbf{(NC3) Self-Duality:} The classifier's weights align with the class means, expressed as \( \frac{\mathbf{W}^\top}{\|\mathbf{W}\|_F} - \frac{\mathbf{\tilde{M}}}{\|\mathbf{\tilde{M}}\|_F} \to 0 \).

    \item \textbf{(NC4) Nearest Class-Center (NCC):} The classifier effectively predicts by assigning features to the nearest class mean, \( \underset{k}{\text{argmax}} \langle \bm{w}_k, f(x) \rangle \to \underset{k}{\text{argmin}} \|f(x) - \bm{z}_k\|_2 \).
\end{itemize}
\vspace{-0.3cm}

\section{Methods}
\label{sec:methods}


This work proposes \textbf{\textit{BalCAL}}, a method with a confidence-tunable module 
to improve model calibration. This confidence-tunable module, based on a fixed Simplex ETF classifier, is detailed in \Cref{subsec:module}, and a dynamic adjustment mechanism is described in \Cref{subsec:adjust}. 
Besides, we also provide a detailed analysis of why the ETF classifier can help improve the model calibration in \Cref{subsec:ETF_classifier}.


\subsection{Why the Simplex ETF classifier can help model calibration?}
\label{subsec:ETF_classifier}



\begin{figure}
    \centering
    \includegraphics[width=0.9\linewidth]{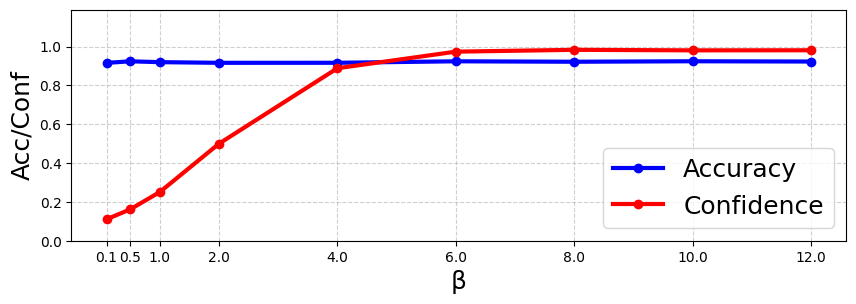}
    \caption{Impact of different \(\beta\) on the output confidence of ETF classifiers on CIFAR-10.}
    \label{figure beta}
    \vspace{-0.5cm}
\end{figure}

The neural collapse phenomenon, observed during the later stages of training, reveals that the features of the final layer and the classifier weights often converge to a Simplex Equiangular Tight Frame (ETF) structure. This structure presents a compelling opportunity to improve model calibration. In this section, we provide a detailed analysis of the relationship between a Simplex ETF classifier and model calibration by presenting the following theorem.
\vspace{-0.5pt}
\begin{theorem}
\label{theorem 1}
In the context of a Simplex ETF initialized as a fixed classifier, the output confidence \( \hat{p}_i\) is related to both the scaling factor \( \beta \) and the number of classes \( K \). Specifically, we have the following relationship:

\begin{equation}
\hat{p}_i \propto \exp\left( \beta \sqrt{\frac{K}{K-1}} \sigma_i \right),
\end{equation}
where  \( \sigma_i \!=\! \bm{z} \hat{\bm{m}}_i\) denotes the score of sample about the $i$-th class and $\hat{\bm{m}}_i$ is the $i$-th columns of the matrix $\mathbf{U} \left( \mathbf{I} - \frac{1}{K} \mathbf{1}_K \mathbf{1}_K^\top \right)$. The difference between \( \hat{p}_i \) and \( p_j \) follows an exponential relationship, which is adjusted by the factor \( \beta \sqrt{\frac{K}{K-1}} \). This can be expressed as:
\begin{equation}
\frac{\hat p_i}{p_j} = \exp\left( \beta \sqrt{\frac{K}{K-1}} (\sigma_i - \sigma_j) \right).
\end{equation}
This equation indicates that the probability ratio between classes \( i \) and \( j \) are governed by the difference in $(\sigma_i - \sigma_j)$, which is further scaled by the factor \( \beta \sqrt{\frac{K}{K-1}} \). It is important to note that while the absolute values of \(\hat p_i \) and \( p_j \) may not be directly affected by this transformation, the relative relationship between these probabilities is indeed adjusted. This highlights the central role of the differences \( \sigma_i - \sigma_j \) in determining the final output, while the scaling factor ensures that this relationship is calibrated appropriately according to the class structure.
\end{theorem}
\vspace{-0.1cm}
We provide the complete proof of Theorem \ref{theorem 1} in Appendix \Cref{proof1}. Theorem~\ref{theorem 1} shows that if the scaling factor $\beta$ increases, the model confidence also increases. This shows that $\beta$ is a parameter that controls the model confidence of ETF classifiers.
Besides, we empirically show the impact of \(\beta\) on the model confidence of an ETF classifier on CIFAR-10 in Fig. \ref{figure beta}, which is in line with Theorem~\ref{theorem 1}. Importantly, it can be observed that increasing \(\beta\) leads to higher model confidence but has little impact on the model accuracy. 

\subsection{A confidence-tunable module}

\label{subsec:module}


Based on previous analysis, we propose a confidence-tunable module built upon a fixed Simplex ETF classifier. This module complements the standard learnable classifier, aiming to regulate prediction confidence dynamically. The pipeline of our proposed approach is illustrated in Figure \ref{fig:overall}.

Specifically, an image \(x  \) can be fed in  a shared encoder to generate the feature vector \( \bm{z} = f(x; \bm{\theta}) \in \mathbb{R}^{ d} \). Given the standard learnable classifier \( \mathbf{W} \in \mathbb{R}^{ d \times K}\), we have the $K$-dimensional probability vector, denoted as $\bm{p}^{Sta} = \text{softmax}(\bm{z}\mathbf{W})$. While the standard classifier predictions often suffer from overconfidence or underconfidence, the fixed ETF classifier introduces a mechanism to balance this behavior. However, directly appending the fixed ETF classifier to the feature extractor could create conflicts due to discrepancies in feature distributions. Therefore, we design the confidence-tunable module by consisting of an adapter \( A(\cdot) \in \mathbb{R}^d \to \mathbb{R}^d\) parameterized with \( \bm{\phi}\) and the fixed ETF classifier \( \mathbf{M} \) defined in \eqref{ETF_classifier}. This configuration ensures smooth integration of the two classifiers.

Taking the feature vector $\bm{z} $ as the input, the confidence-tunable module produces the $K$-dimensional probability vector $\bm{p}^{ETF} = \text{softmax}(\bm{z}'\mathbf{M})$ where $\bm{z}'=A(\bm{z};\bm{\phi})$. Therefore, for the same input, we have two probability vectors. Respectively, the two classification losses for all samples can be represented as follows:

\begin{equation} 
\mathcal{L_{\text{Sta}}} = -\frac{1}{N}  \sum_{i=1}^{N}\sum_{k=1}^{K} y_{i}^{k} \log(p_{i,k}^{Sta}),
\label{eq:loss_learnable}
\end{equation}

\begin{equation} 
    \mathcal{L_{\text{ETF}}} = -\frac{1}{N}  \sum_{i=1}^{N}\sum_{k=1}^{K} y_{i}^{k} \log(p_{i,k}^{ETF}).
\label{eq:loss_cal}
\end{equation}

Now the parameters $\{\bm{\theta}, \mathbf{W},\bm{\phi}\}$ can be optimized by
minimizing the two losses combined with parameter $\gamma$:

\begin{equation}
\min_{\bm{\theta}, \mathbf{W},\bm{\phi}} \mathcal{L_{\text{total}}} = \gamma \min_{\bm{\theta},\mathbf{W}} \mathcal{L_{\text{Sta}}} + (1 - \gamma) \min_{\bm{\theta},\bm{\phi}} \mathcal{L_{\text{ETF}}}.
\label{equation:loss_total}
\end{equation}

In our method, $\beta$ controls the confidence of the ETF classifier while $\gamma$ controls the calibration intensity of the ETF classifier. Empirically, we find that $\gamma$ has a more significant impact on the confidence of the combined classifier than $\beta$. Moreover, we find that
the calibration intensity needs to vary during the training process of the model as in the early stage of training, the model should focus more on improving its accuracy with less regularization on confidence while in the late stage, more calibration intensity may be needed.
To dynamically control calibration intensity during training, we propose an adaptive mechanism detailed in Section \ref{subsec:adjust}.
In terms of $\beta$, we set $\beta=1$ to the ETF classifier relatively underconfident for models prone to overconfidence (e.g., typical DNNs) and $\beta=K$ to enhance the ETF classifier's confidence for underconfident models (e.g., those trained with Mixup).

\subsection{Dynamic adjustment mechanism}
\label{subsec:adjust}
{Once we introduce the confidence-tunable module, we need to solve one key problem during the training process, where the parameter \(\gamma\) plays a key role, which can balance the classification loss from the ETF classifier and loss from the standard classifier. Therefore, how to design the balanced factor $\gamma$ is our focus in this section. Considering that the model calibration methods aim to match predicted confidence with actual accuracy, we utilize the training dataset to evaluate the model confidence \(\text{conf}_t(D, \gamma )\) and accuracy \(\text{acc}_t(D, \gamma)\) at the \(t\)-th  training epoch, whose details are described below. By the relationship between model confidence and accuracy at the \(t\)-th training epoch, we can determine the calibration error and design the balanced factor $\gamma^{t+1}$ at the \(t+1\)-th training epoch as follows:
\begin{equation}
\gamma_{t+1} = \min_{\gamma} \left| \text{conf}_t(D, \gamma) - \delta \cdot \text{acc}_t(D, \gamma) \right|,
\label{equation:gamma}
\end{equation}
where we use binary search for $\gamma_{t+1}$. The hyperparameter $\delta$ is introduced to mitigate the overestimation of training-set accuracy caused by overfitting, by scaling accuracy to mimic unseen-data behavior. This design prevents model overconfidence and helps identify calibration errors between predicted confidence and actual accuracy. Through dynamic $\gamma_t$ adjustment, BalCAL automatically detects calibration errors and balances classifiers, achieving improved calibration with minimal manual intervention.
} 

To compute the confidence and accuracy at the $t$-th training epoch, we only need to fuse the probability vectors produced by two classifiers for each training sample in the mini-batch as follows:
\begin{equation}
\bm{p}^{fused} = \gamma \bm{p}^{Sta} + (1 - \gamma) \bm{p}^{ETF}.
\label{equation:p_fused}
\end{equation}
Then we can compute the accuracy \(\text{acc}_t(D , \gamma)\)  and confidence \(\text{conf}_t(D, \gamma )\) on the training dataset by averaging all mini-batches at the \(t\)-th  training epoch. Notably, during the test stage, we utilize the $\gamma$  accompanied by the best saved model to fuse the probability vectors. The mixing strategy ensures that the predicted confidence accurately aligns with the actual accuracy for improving model reliability and stability in real-world applications. We summarize our proposed method in \cref{Algorithm}.


\begin{algorithm}[t]
    \renewcommand{\algorithmicrequire}{\textbf{Input:}}
    \renewcommand{\algorithmicensure}{\textbf{Output:}}
    \caption{Overall Algorithm}
    \label{alg:balcal_training}
    \begin{algorithmic}[1]
        \REQUIRE Training dataset $D$, feature extractor $f$ with parameter $\bm{\theta}$, learnable classifier $\mathbf{W}$, Adapter $A$ with parameter $\bm{\epsilon}$, fixed ETF classifier $\mathbf{M}$,  hyper-parameters $\delta$;
        \ENSURE Trained model parameters and balanced factor $\gamma$;
        
        \STATE Initialize model parameters randomly, and set $\gamma_{1}=0.5$
        
        \FOR{$t = 1, 2, \ldots, T$}
            \FOR{each mini-batch $\{x, y\} \in D$}
            
                \STATE Calculate and save probabilities  $\bm{p}^{Sta} $ and $\bm{p}^{ETF}$;
                
                 \STATE Compute two classification losses $\mathcal{L_{\text{Sta}}}$ with \cref{eq:loss_learnable} and $\mathcal{L_{\text{ETF}}}$  with \cref{eq:loss_cal};

                \STATE Update \{$\bm{\theta}$, $\bm{\mathbf{W}}$, $\bm{\epsilon}$\} by minimizing $\gamma_t\mathcal{L_{\text{Sta}}} + (1 - \gamma_t)\mathcal{L_{\text{ETF}}}$ with \cref{equation:loss_total};
            \ENDFOR
            \STATE Use binary search to determine the optimal $\gamma_{t+1}$ by minimizing $\left| conf_t(D) - \delta \cdot acc_t(D ) \right|$ with \cref{equation:gamma};
        \ENDFOR
    \end{algorithmic}
    \label{Algorithm}
\end{algorithm}

\section{Experiments}
\label{sec:exp}

\begin{table*}[htbp]
  \centering
    \resizebox{0.9\textwidth}{!}{
    \begin{tabular}{c|l l l|l l l|l l l|c}

    \toprule
          \multirow{2}[2]{*}{\textbf{Methods}} & \multicolumn{3}{c|}{\textbf{CIFAR-10}} & \multicolumn{3}{c|}{\textbf{CIFAR-100}} & \multicolumn{3}{c|}{\textbf{SVHN}} & \multirow{2}[2]{*}{\textbf{ECE Avg}} \\
\cmidrule{2-10}          & ACC↑  & ECE↓  & AECE↓ & ACC↑  & ECE↓  & AECE↓ & ACC↑  & ECE↓  & AECE↓ \\
    \midrule
    Vanilla & 91.34  & 6.38  & 6.38  & 70.79  & 21.63  & 21.63  & 95.12  & 3.32  & 3.31  & 10.44 \\
    ACLS\citep{park2023acls}  & 93.12  & 5.26  & 5.26  & 72.36  & 13.08  & 13.08  & 95.33  & 3.20  & 3.19  & 7.18 \\
    LS\citep{szegedy2016rethinking}    & 92.12  & 2.79  & 4.26  & 73.58  & 6.03  & 6.03  & \textbf{95.50}  & 3.77  & 3.93  & 4.20 \\
    Mixup\citep{thulasidasan2019mixup} & \textbf{94.71}  & 2.56  & 3.08  & \textbf{77.05}  & 4.51  & 4.58  & 94.82  & 3.75  & 3.57  & 3.61 \\
    MIT\citep{wang2023pitfall}   & 94.27  & 2.18  & 2.12  & 73.38  & 12.31  & 12.31  & 93.69  & 0.78  & 0.68  & 5.09 \\
    FL\citep{mukhoti2020calibrating}    & 92.11  & 1.99  & 1.97  & 71.54  & 14.26  & 14.25  & 94.79  & 1.24  & 1.26  & 5.83\\
    FLSD\citep{mukhoti2020calibrating}  & 92.42  & 1.72  & 1.68  & 71.19  & 14.20  & 14.20  & 94.60  & 1.03  & 0.92  & 5.65\\
    \midrule
    CPC\citep{Cao2019}   & 91.56  & 6.28  & 6.25  & 73.33  & 12.53  & 12.64  & 94.73  & 1.61  & 1.45  & 6.81\\
    PLP\citep{wang2024calibration}   & 92.69  & 4.89  & 4.85  & 72.55  & 13.74  & 13.73  & 95.07  & 3.49  & 3.49  & 7.37\\
    MMCE\citep{kumar2018trainable}  & 90.04  & 3.60  & 3.43  & 68.77  & 17.18  & 17.18  & 93.99  & 2.30  & 2.31  & 7.69\\
    TST$^*$\citep{jordahn2024decoupling}   & 92.59\textsubscript{±0.02} & 2.18\textsubscript{±0.353} &   \phantom{ }  -    & 71.26\textsubscript{±0.06}  & 7.07\textsubscript{±0.084} &  \phantom{ }  -    & 95.1\textsubscript{±0.01} & 0.43\textsubscript{±0.016} &   \phantom{ }  -    & 3.23  \\
    VTST$^*$\citep{jordahn2024decoupling}  & 91.09\textsubscript{±0.06} & 1.52\textsubscript{±0.144} &   \phantom{ }  -    & 69.18\textsubscript{±0.09} & 7.34\textsubscript{±0.291} &  \phantom{ }  -    & 94.73\textsubscript{±0.02} & 0.48\textsubscript{±0.067} &   \phantom{ }  -    & 3.11 \\
    \rowcolor{gray!20} 
    Ours  & 92.23\textsubscript{±0.28} & \textbf{0.76}\textsubscript{±0.26} & \textbf{0.82}\textsubscript{±0.26} & 73.21\textsubscript{±0.28} & \textbf{4.21}\textsubscript{±0.14} & \textbf{4.13}\textsubscript{±0.24} & 95.47\textsubscript{±0.23} & \textbf{0.24}\textsubscript{±0.11} & \textbf{0.33}\textsubscript{±0.12} & \textbf{1.77} \\
    \hline
    \end{tabular}
    }
    \caption{Comparison results of our proposed method against state-of-the-art methods for model calibration on CIFAR-10, CIFAR-100, and SVHN. $^*$ denotes the results are from original papers, while the remaining results correspond to the baselines that we reproduced.}
  \label{tab:tate-of-the-art}
\end{table*}


\begin{table}[!htbp]
  \label{table 2}
  \centering
  \resizebox{0.85\columnwidth}{!}{
    \begin{tabular}{c|ccc|ccc}
    \toprule
    \multirow{2}[2]{*}{\textbf{Methods}} & \multicolumn{3}{c|}{\textbf{Resnet-50}} & \multicolumn{3}{c}{\textbf{Resnet-101}} \\
          \cmidrule{2-7} & ACC↑   & ECE↓   & AECE↓  & ACC↑   & ECE↓   & AECE↓ \\
    \midrule
    Vanilla & 60.53  & 4.38  & 4.37  & 62.05  & 2.50  & 2.44  \\
    LS    & 63.10  & 1.59  & 1.71  & 63.12  & 1.93  & 6.25  \\
    mixup & \textbf{65.88} & 3.85  & 3.84  & 66.56  & 3.98  & 3.86  \\
    MIT   & 64.27  & \underline{1.45}  & \textbf{1.32} & \underline{66.81}  & 1.58  & \underline{1.64}  \\
    FL    & 62.82  & 2.07  & 2.05  & 62.37  & 2.57  & 7.76  \\
    FLSD  & 63.09  & 1.84  & 1.84  & 63.09  & \underline{1.39}  & 1.65  \\
    \midrule
    TST   & 59.96  & 6.18  & 6.18  & 61.49  & 6.01  & 6.01  \\
    VTST  & 59.46  & 5.75  & 5.75  & 60.85  & 2.91  & 2.88  \\
    \rowcolor{gray!20} 
    \textbf{Ours}  & \underline{65.22}  & \textbf{1.40} & \underline{1.47}  & \textbf{66.90} & \textbf{1.35} & \textbf{1.48} \\
    \bottomrule
    \end{tabular}%
  }
  \caption{Calibration performance on Tiny-ImageNet with Resnet-50 and Resnet-101.}
  \label{tab:backbone}
\end{table}%
\paragraph{Datasets and Baselines.}  
We evaluate model calibration on CIFAR-10/100 \citep{krizhevsky2009learning}, SVHN \citep{netzer2011reading}, and Tiny-ImageNet \citep{le2015tiny}, along with distribution-shifted datasets CIFAR-10-C/100-C \citep{hendrycks2019benchmarking}. ImageNet results and additional experiments are in Appendix \ref{MoreResults}.  
We compare BalCAL with various calibration methods: (1) Vanilla model trained with CE loss; (2) Implicit methods: ACLS \citep{park2023acls}, LS \citep{szegedy2016rethinking}, Mixup \citep{thulasidasan2019mixup}, MIT \citep{wang2023pitfall}, FL/FLSD \citep{mukhoti2020calibrating}; (3) Explicit methods: CPC \citep{Cao2019}, PLP \citep{wang2024calibration}, MMCE \citep{kumar2018trainable}, and TST/VTST \citep{jordahn2024decoupling}. When available, we reproduce results using official code and hyperparameters. Details are in Appendix \ref{Implementation}.  

\paragraph{Training Details and Metrics.}  
Following \citep{jordahn2024decoupling}, we use WRN-28-10 \citep{zagoruyko2016wide} for CIFAR-10/100 and SVHN, and ResNet50/101 \citep{He_2016_CVPR} for Tiny-ImageNet. Additional architectures (e.g., ViT, DenseNet) are in Appendix \ref{MoreResults}.  
Calibration is evaluated using accuracy, expected calibration error (ECE) \citep{guo2017calibration}, and adaptive ECE (AECE) \citep{nguyen2015posterior}. OOD detection is assessed via AUROC \citep{bradley1997use} and FPR95. For further details, see Appendix \ref{Evaluation protocols.}.

\begin{figure*}
    \centering
    \includegraphics[width=1\linewidth]{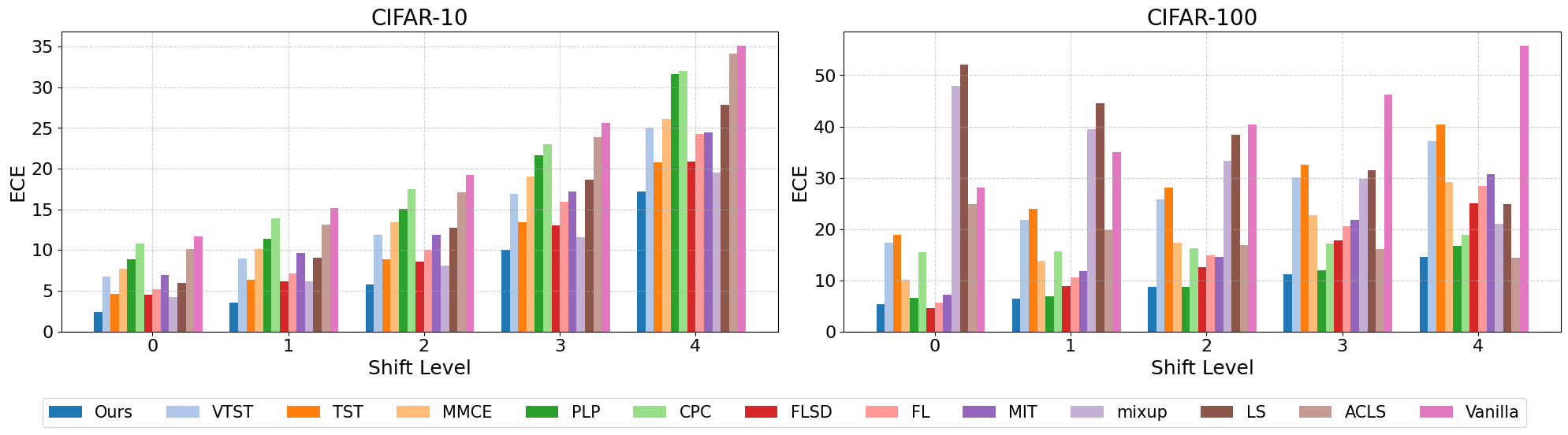}
    \caption{{Calibration performance under distribution shifts of various methods on CIFAR-10 (left) and CIFAR-100 (right). Here, x-axis denotes the shift levels (the larger the number, the greater the shift degree) and y-axis is the ECE (smaller is better).}}
    \label{fig:shift}
    \vspace{-0.5cm}
\end{figure*}
\vspace{-0.15cm}
\subsection{Results}
\paragraph{Comparison with other methods.}
We present comparison results of our proposed method against state-of-the-art methods for model calibration on three standard datasets, as illustrated in \Cref{tab:tate-of-the-art}. We can find that our method significantly enhances calibration performance across all datasets, achieving superior results on both ECE and AECE, especially on the average of ECE. Compared with competed TST, our method achieves 1.42\% and 2.86\% ECE improvement on CIFAR-10 and CIFAR-100, respectively. 
While some methods achieve higher accuracy, their calibration performance is often suboptimal, without addressing the balance between predicted confidence and accuracy very well, which however is the focus in model calibration. Different from them, our proposed BalCAL not only achieves the desired calibration performance but also produces the accepted classification accuracy, which surpasses that of the vanilla model always, for example. This demonstrates the effectiveness of ours for introducing a fixed ETF classifier to correct the output of learnable classifier.


{Following \citep{liu2022devil}}, we further conduct additional experiments on Tiny-ImageNet, which presents serious challenges compared with CIFAR-10/100 and SVHN. We adopt ResNet-50 and ResNet-101 as the backbones. \Cref{tab:backbone} shows that ours performs consistently well on Tiny-ImageNet with two backbones. These experiments indicate the ability of ours to handle challenging datasets with different architectures.
\vspace{-0.3cm}
\paragraph{Underconfident problem.} 
\label{underconfident problem}

\begin{figure}
    \centering
    \includegraphics[width=0.6\linewidth]{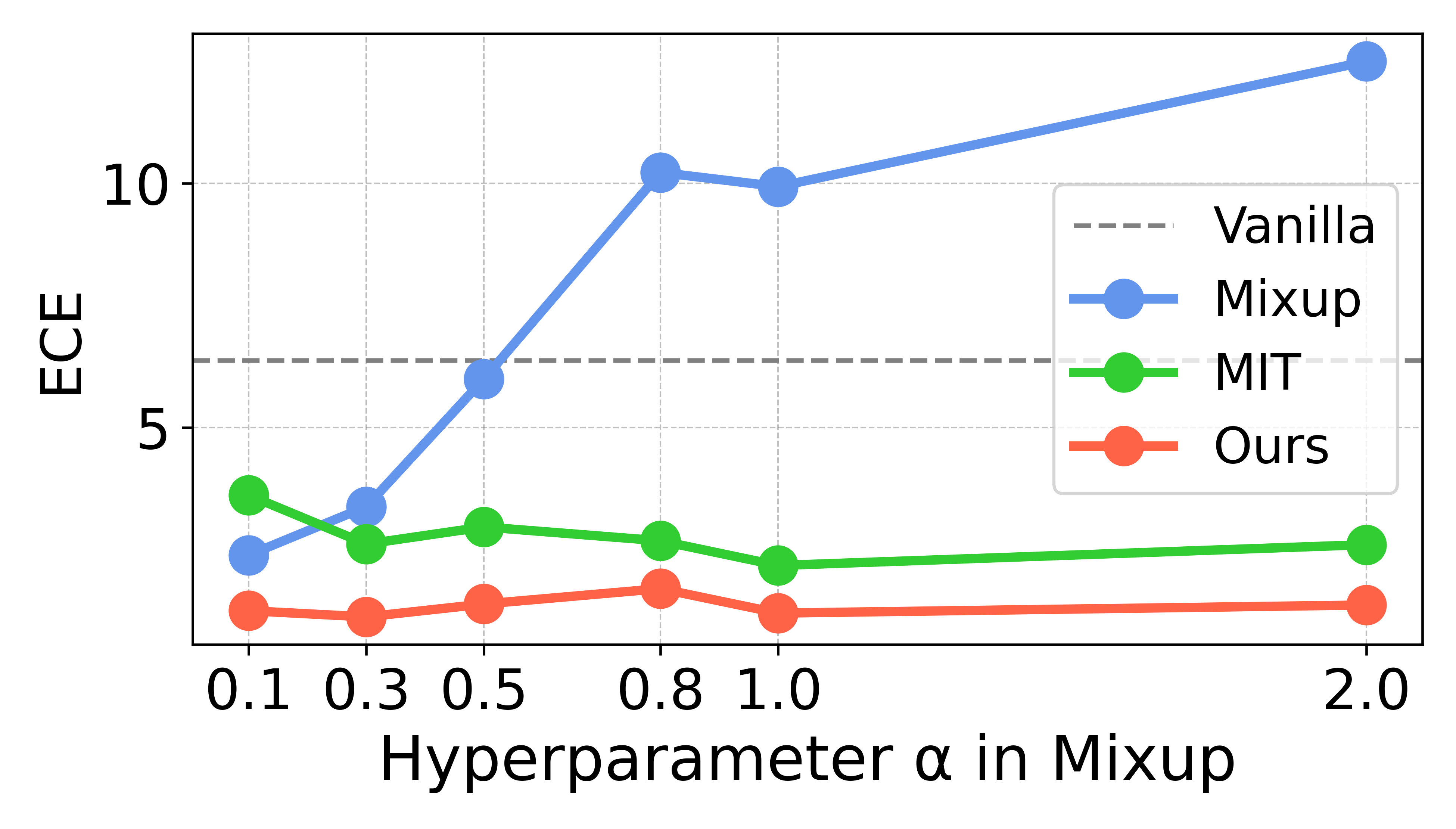}
    \caption{Calibration sensitivity of Mixup, MIT, and Mixup+Ours with combination ratios $\alpha$ on CIFAR-10.}
    \label{fig:mixup}
    \vspace{-0.5cm}
\end{figure}

As discussed above, Mixup improves model calibration but is sensitive to interpolation levels, which sometimes can even lead to underconfidence. Specifically, the combination ratio on Mixup is sampled from Beta distribution, denoted as $\text{Beta}(\alpha,\alpha)$; see mode details about Mixup in Appendix \ref{MoreResults}. Recent work \cite{wang2023pitfall} observed that a larger $\alpha$ used in Mixup may cause underconfidence issue, which makes the selection of $\alpha$ a difficulty. They thus proposed a specialized strategy named mixup inference in training (MIT). To explore whether ours can address the underconfident problem by Mixup, we compare the ECE of vanilla, Mixup, MIT, and Mixup+ours with varying hyperparameter $\alpha$ in \cref{fig:mixup}.  We can find that Mixup is sensitive to $\alpha$, where  the model gradually suffers from underconfidence when $\alpha>0.1$ and we thus set $\beta=K$ for $\alpha>0.1$ in ETF classifier; please see the specific accuracy and confidence values in Appendix \ref{MoreResults}. Both MIT and ours can mitigate the pitfall of mixup for calibration. Without specialized design, ours can even perform better ECE than MIT, proving the effectiveness of ours.






\vspace{-0.3cm}
\paragraph{Robustness analytical.}
To evaluate the robustness of our proposed BalCAL under corruptions and perturbations for model calibration, we further report the comparison experiments with several robustness datasets. We train on CIFAR-10/100 and test on CIFAR-10-C and CIFAR-100-C \citep{hendrycks2019benchmarking}, respectively. 
\cref{fig:shift} shows the ECE performance of BalCAl against various model calibration methods with different shift levels. {We find that ours achieves more desired ECE values than our baselines when facing higher shift levels. Especially on the classification task of CIFAR-100, the ECE of our method is much lower than the other methods when the shift level is high. These indicate the stability and robustness of our proposed confidence-tunable module for model calibration in difficult scenarios.  Besides, we also provide experiments about the OOD detection, where we train a model on one dataset (e.g., CIFAR-10) and test it on other datasets (e.g., CIFAR-100 and SVHN). As shown in \Cref{tab:ood1}, our related baseline TST and VTST perform generally
 worse than the vanilla model but ours achieves best AUROC and FPR95. It proves that, instead of harming performance on OOD tasks like some model calibration methods, ours can surprisingly improve the generalization ability. The possible reason might be that we introduce two classifiers, which can not only correct the model confidence but also improve the generalization ability. }




\begin{table}
  \centering
  \resizebox{\linewidth}{!}{
    \begin{tabular}{c|llllll}
    \toprule
          \multirow{2}[2]{*}{\textbf{Methods}} & \multicolumn{2}{c}{\textbf{CIFAR-10}}   & \multicolumn{2}{c}{\textbf{CIFAR-100}}  & \multicolumn{2}{c}{\textbf{SVHN}} \\
         \cmidrule{2-7} & AUROC↑ & FPR95↓ & AUROC↑ & FPR95↓ & AUROC↑ & FPR95↓ \\
    \midrule
    Vanilla & 88.56  & 70.98  & 67.05  & 89.26  & 93.46  & 45.92\\
    PLP\citep{wang2024calibration}   & 86.80  & 73.62  & 71.90  & 83.12  & 93.64  & 44.78 \\
    TST\citep{jordahn2024decoupling}   & 88.47  & 67.28  & 71.27  & 85.60  & 93.73  & 42.15\\
    VTST\citep{jordahn2024decoupling}  & 83.47  & 72.40  & 73.47  & 84.93  & 83.50  & 60.91\\
        \rowcolor{gray!20} 
    \textbf{Ours} & \textbf{89.89} & \textbf{64.15}& \textbf{81.26}& \textbf{76.03} & \textbf{94.98} & \textbf{34.59}\\
    \bottomrule
    \end{tabular}%
  \label{tab:ood1}%
 }
 \caption{Performance comparison of OOD detection. {Dataset name denotes the training dataset, where we evaluate the model on two datasets other than the training set and report the averaged performance.}}
\end{table}%

\subsection{Ablation study}

\paragraph{Effect of our components.} 

Here, we evaluate several variants to validate the effect of our proposed adapter, the fixed ETF classifier, and the dynamic adjustment mechanism. As shown in \Cref{tab:variants}, replacing the learnable classifier with a fixed ETF classifier, i.e., using only a classifier, we consider two settings: $\beta=1$ and $\beta=K$. 
We can see that, if we utilize the fixed ETF classifier with a fixed $\beta$ and only train the encoder, it performs worse or is similar to the vanilla method. It indicates that only a fixed ETF classifier can not used to solve the model calibration problem and the performance is sensitive to $\beta$, where we provide its additional results with more different $\beta$ values in Appendix \ref{MoreResults}. Besides, considering that we introduce a dynamic adjustment mechanism to search the balance factor $\gamma$, we can also use the mechanism to build a dynamic $\beta$ during the training process, which can be implemented by replacing $\gamma$ with $\beta$ in \eqref{equation:gamma}. We can find that when tuning the scaling factor $\beta$ with our proposed mechanism, only a fixed ETF classifier can produce the second-best ECE values, proving the effectiveness of our proposed dynamic adjustment mechanism. However, it is still inferior to ours where we combine the learnable classifier and ETF classifier, proving the importance of their combination. Besides, removing the adapter in ours can result in a decline in performance, which is reasonable since an adapter can reduce the conflicts between the fixed ETF classifier and the encoder.

\begin{table}
  \centering
  \resizebox{\columnwidth}{!}{
    \begin{tabular}{c|c|c|cccccc}
    \toprule
    \multicolumn{1}{c|}{\textbf{Learnable}} & \multicolumn{1}{c|}{\textbf{ETF}} & \multicolumn{1}{c|}{\multirow{2}[4]{*}{\textbf{Adapter}}}  & \multicolumn{2}{c}{\textbf{CIFAR-10}} & \multicolumn{2}{c}{\textbf{CIFAR-100}} & \multicolumn{2}{c}{\textbf{SVHN}} \\
\cmidrule{4-9}    \multicolumn{1}{c|}{\textbf{Classifier}} & \multicolumn{1}{c|}{\textbf{Classifier}} & \multicolumn{1}{c|}{}  & ACC↑  & ECE↓  & ACC↑  & ECE↓  & ACC↑  & ECE↓ \\
    \midrule
      $\surd$    &       &       & 91.33 & 6.38(1.12) & 70.78 & 21.63(\textbf{3.10}) & 95.12 & 3.32(\underline{0.72}) \\
          &   $\beta$=1    &       & 91.99 & 66.91(\underline{0.77}) & 72.22 & 68.96(6.31) & \textbf{95.89} & 70.76(1.27) \\ 
          &   $\beta$=K    &      & \underline{92.15} & 7.85(1.81) & 71.98 & 28.15(5.32) & 95.32 & 3.86(1.00) \\
          & \text{dynamic} $\beta$   &       & 91.9  & \underline{0.87}(0.87) & 71.53 & \underline{5.78}(5.78) & \textbf{95.80}  & \underline{0.80}(0.92) \\
      $\surd$    &    $\beta$=1  &          & 92.14 & 1.72(1.03) & \underline{72.86} & 9.34(8.79) & 95.45 & 0.90(0.83) \\
      \rowcolor{gray!20} 
       $\surd$   &    $\beta$=1, \text{dynamic} $\gamma$   &    $\surd$   & \textbf{92.23} & \textbf{0.76}(\textbf{0.69}) & \textbf{73.21} & \textbf{4.21}(\underline{3.97}) & 95.47 & \textbf{0.24}(\textbf{0.59}) \\
    \bottomrule
    \end{tabular}
  }
  \caption{Our method and its variants. Numbers in parentheses represent the results obtained using a TS post-hoc method \citep{guo2017calibration}.}
  \label{tab:variants}
\end{table}%





\begin{figure}
    \centering
    \includegraphics[width=0.8\linewidth]{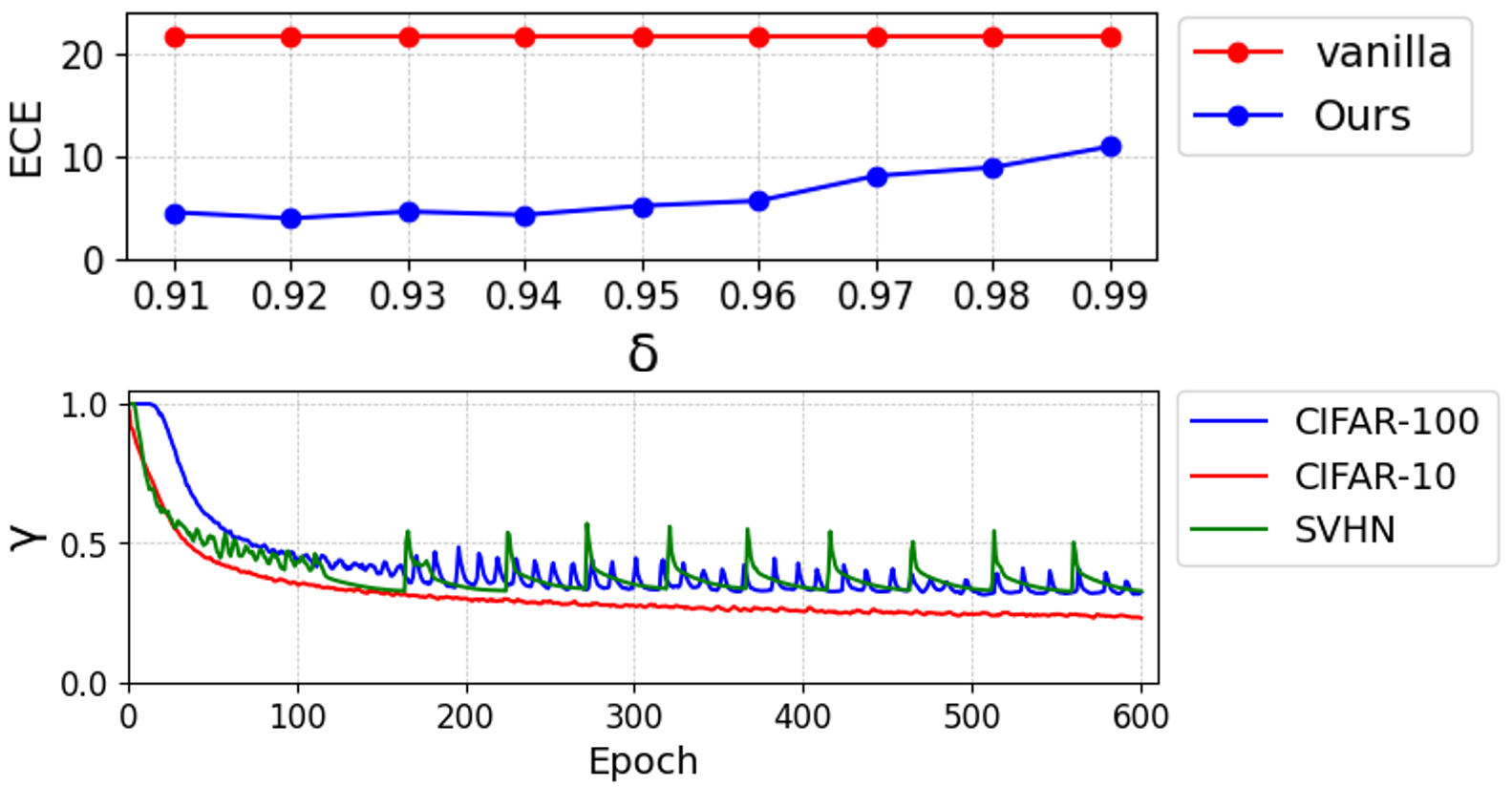}
    \caption{Above: impact of different \(\delta\) values on the calibration performance of BalCAL on CIFAR-100. Below: dynamic variation of \(\gamma\) across epochs in WRN-28-10 under \(\delta\).}
    \label{fig:delta}
    \vspace{-0.6cm}
\end{figure}

\vspace{-0.5cm}
\paragraph{Cautious parameter $\delta$.}
The cautious hyperparameter \(\delta\) in BalCAl encourages the model to maintain suitable uncertainty, which helps identify the calibration error between predicted confidence and accuracy, and adaptively determine the balanced factor \(\gamma\). We recommend setting \(\delta\) within the range of \([0.91, 0.99]\) based on empirical results. As shown in the upper part of \cref{fig:delta}, \(\delta\) within this range consistently leads to significant calibration improvements.  
As shown in the lower part of \cref{fig:delta}, we illustrate the dynamic variation of the balanced factor \(\gamma\) across epochs under the influence of \(\delta\). We can see that, at the beginning of training, the standard classifier plays a more important role since the attention of the model is mainly on the learning of the standard classifier. At the early training stage, the importance of ETF classifier gradually increases; then the change about $\gamma$ gradually becomes smaller.


\subsection{Visualization Results and Analysis}
\paragraph{t-SNE visualization.}
\label{par:tsne}
\begin{figure}
    \centering
    \includegraphics[width=1.0\linewidth]{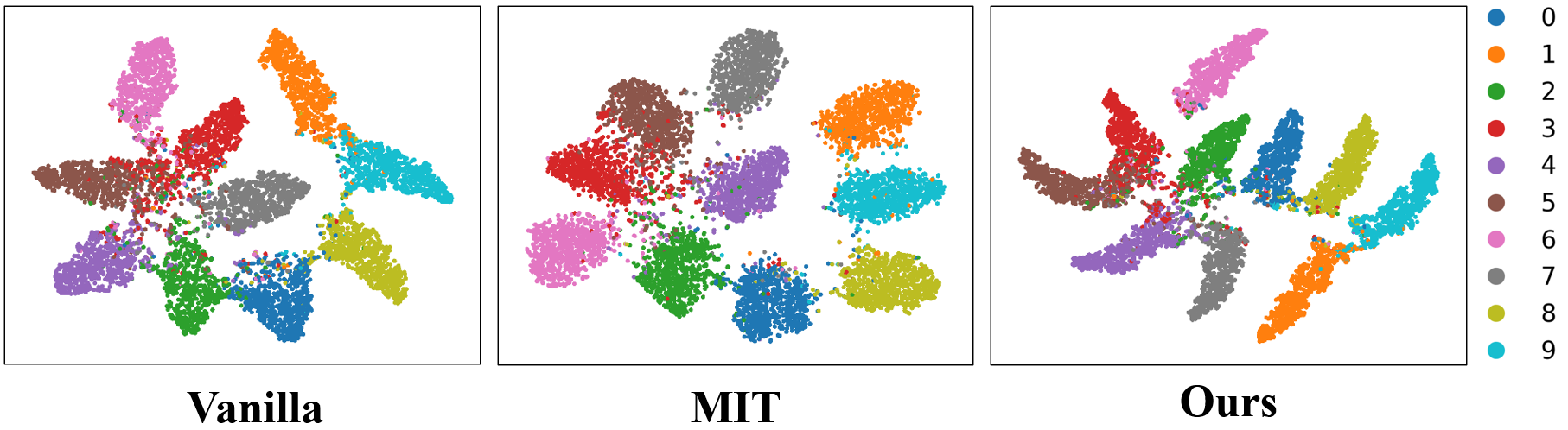}
    \caption{t-SNE visualizations of representations on CIFAR-10.}
    \label{fig:tsne}
\end{figure}
    We present t-SNE visualizations of the last-layer representations from Vanilla, MIT, and our proposed method on CIFAR-10 in \cref{fig:tsne}. We observe that ours exhibits more clustered intra-class and discriminative inter-class distances, demonstrating its effectiveness in enhancing representation learning and potentially balancing confidence and accuracy to address overconfidence or underconfidence issues in model calibration.
\vspace{-0.4cm}
\paragraph{CAMs visualization.}
\begin{figure}
    \centering
    \resizebox{0.8\columnwidth}{!}{
    \includegraphics[width=1.0\linewidth]{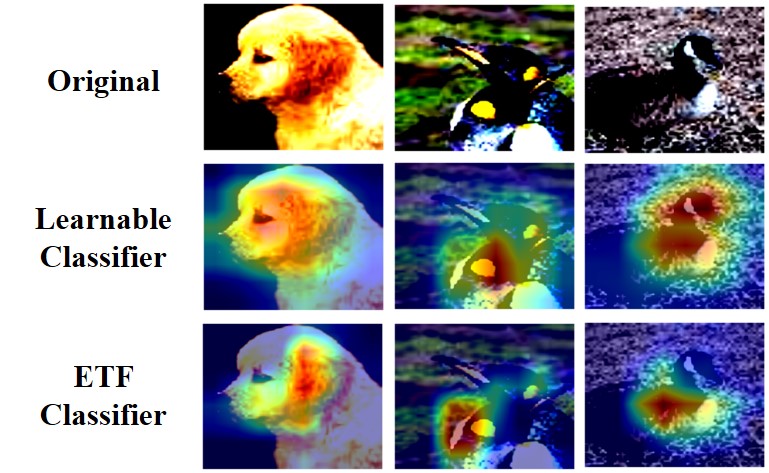}
    }
    \caption{Class activation maps of two classifiers on our method.}
    \label{fig:cam}
    \vspace{-0.7cm}
\end{figure}
We visualize Class Activation Maps for learnable and ETF classifiers in \cref{fig:cam}, where the learnable classifier captures broader regions, while the ETF classifier focuses on localized details. This underscores the ETF classifier’s role in refining predictions, making its combination with the learnable classifier beneficial for enhancing perceptual vision and reducing confidence errors.

\section{Conclusion}
\label{sec:conclusion}
In this paper, we propose a novel method, denoted as \textbf{BalCAL}, which balances two classifiers to address the overconfidence and also underconfidence issues in model calibration. By integrating a fixed Simplex ETF classifier inspired by the NC phenomenon, with a standard learnable classifier, we introduce a dynamic adjustment method to adaptively regulate output confidence without sacrificing model accuracy. Extensive experiments show that our method not only improves model calibration but also enhances robustness under corruptions across various settings. 

\paragraph{Acknowledgements.}
 This work of Dandan Guo was supported by the National Natural Science Foundation of China (NSFC) under Grant 62306125. 


{
    \small
    \bibliographystyle{ieeenat_fullname}
    \bibliography{main}

\begin{thebibliography}{54}
\providecommand{\natexlab}[1]{#1}
\providecommand{\url}[1]{\texttt{#1}}
\expandafter\ifx\csname urlstyle\endcsname\relax
  \providecommand{\doi}[1]{doi: #1}\else
  \providecommand{\doi}{doi: \begingroup \urlstyle{rm}\Url}\fi

\bibitem[Ahsan et~al.(2022)Ahsan, Luna, and Siddique]{ahsan2022machine}
Md~Manjurul Ahsan, Shahana~Akter Luna, and Zahed Siddique.
\newblock Machine-learning-based disease diagnosis: A comprehensive review.
\newblock In \emph{Healthcare}, page 541. MDPI, 2022.

\bibitem[Bradley(1997)]{bradley1997use}
Andrew~P. Bradley.
\newblock The use of the area under the roc curve in the evaluation of machine learning algorithms.
\newblock \emph{Pattern Recognition}, 30\penalty0 (7):\penalty0 1145--1159, 1997.

\bibitem[Cao et~al.(2019)Cao, Wei, Gaidon, Arechiga, and Ma]{Cao2019}
Kaidi Cao, Colin Wei, Adrien Gaidon, Nikos Arechiga, and Tengyu Ma.
\newblock Learning imbalanced datasets with label-distribution-aware margin loss.
\newblock In \emph{Advances in Neural Information Processing Systems (NeurIPS)}. Curran Associates, Inc., 2019.

\bibitem[Chen and Huang(2017)]{chen2017end}
Zhilu Chen and Xinming Huang.
\newblock End-to-end learning for lane keeping of self-driving cars.
\newblock In \emph{2017 IEEE intelligent vehicles symposium (IV)}, pages 1856--1860. IEEE, 2017.

\bibitem[Cheng and Vasconcelos(2022)]{cheng2022calibrating}
Jiacheng Cheng and Nuno Vasconcelos.
\newblock Calibrating deep neural networks by pairwise constraints.
\newblock In \emph{Proceedings of the IEEE/CVF Conference on Computer Vision and Pattern Recognition (CVPR)}, 2022.

\bibitem[Clements et~al.(2020)Clements, Xu, Yousefi, and Efimov]{clements2020sequential}
Jillian~M Clements, Di Xu, Nooshin Yousefi, and Dmitry Efimov.
\newblock Sequential deep learning for credit risk monitoring with tabular financial data.
\newblock \emph{arXiv preprint arXiv:2012.15330}, 2020.

\bibitem[Deng et~al.(2009)Deng, Dong, Socher, Li, Li, and Fei-Fei]{deng2009imagenet}
Jia Deng, Wei Dong, Richard Socher, Li-Jia Li, Kai Li, and Li Fei-Fei.
\newblock Imagenet: A large-scale hierarchical image database.
\newblock In \emph{2009 IEEE conference on computer vision and pattern recognition}, pages 248--255. Ieee, 2009.

\bibitem[Devlin(2018)]{devlin2018bert}
Jacob Devlin.
\newblock Bert: Pre-training of deep bidirectional transformers for language understanding.
\newblock \emph{arXiv preprint arXiv:1810.04805}, 2018.

\bibitem[Dosovitskiy et~al.(2021)Dosovitskiy, Beyer, Kolesnikov, Weissenborn, Zhai, Unterthiner, Dehghani, Minderer, Heigold, Gelly, et~al.]{dosovitskiy2020image}
Alexey Dosovitskiy, Lucas Beyer, Alexander Kolesnikov, Dirk Weissenborn, Xiaohua Zhai, Thomas Unterthiner, Mostafa Dehghani, Matthias Minderer, Georg Heigold, Sylvain Gelly, et~al.
\newblock An image is worth 16x16 words: Transformers for image recognition at scale.
\newblock In \emph{Proceedings of the International Conference on Learning Representations (ICLR)}, 2021.

\bibitem[Gao et~al.(2024)Gao, Zhao, Guo, and Zha]{gao2024distribution}
Jintong Gao, He Zhao, Dandan Guo, and Hongyuan Zha.
\newblock Distribution alignment optimization through neural collapse for long-tailed classification.
\newblock In \emph{Proceedings of the International Conference on Machine Learning (ICML)}, 2024.

\bibitem[Ghosh et~al.(2022)Ghosh, Schaaf, and Gormley]{ghosh2022adafocal}
Arindam Ghosh, Thomas Schaaf, and Matthew~R Gormley.
\newblock Adafocal: Calibration-aware adaptive focal loss.
\newblock In \emph{Proceedings of the 36th Conference on Neural Information Processing Systems (NeurIPS)}, 2022.

\bibitem[Guo et~al.(2017)Guo, Pleiss, Sun, and Weinberger]{guo2017calibration}
Chuan Guo, Geoff Pleiss, Yu Sun, and Kilian~Q Weinberger.
\newblock On calibration of modern neural networks.
\newblock In \emph{Proceedings of the 34th International Conference on Machine Learning (ICML)}, 2017.

\bibitem[Gupta et~al.(2021)Gupta, Anpalagan, Guan, and Khwaja]{gupta2021deep}
Abhishek Gupta, Alagan Anpalagan, Ling Guan, and Ahmed~Shaharyar Khwaja.
\newblock Deep learning for object detection and scene perception in self-driving cars: Survey, challenges, and open issues.
\newblock \emph{Array}, 10:\penalty0 100057, 2021.

\bibitem[He et~al.(2016{\natexlab{a}})He, Zhang, Ren, and Sun]{He_2016_CVPR}
Kaiming He, Xiangyu Zhang, Shaoqing Ren, and Jian Sun.
\newblock Deep residual learning for image recognition.
\newblock In \emph{Proceedings of the IEEE Conference on Computer Vision and Pattern Recognition (CVPR)}, 2016{\natexlab{a}}.

\bibitem[He et~al.(2016{\natexlab{b}})He, Zhang, Ren, and Sun]{he2016deep}
Kaiming He, Xiangyu Zhang, Shaoqing Ren, and Jian Sun.
\newblock Deep residual learning for image recognition.
\newblock In \emph{Proceedings of the IEEE Conference on Computer Vision and Pattern Recognition (CVPR)}, pages 770--778, 2016{\natexlab{b}}.

\bibitem[He et~al.(2017)He, Gkioxari, Doll{\'a}r, and Girshick]{he2017mask}
Kaiming He, Georgia Gkioxari, Piotr Doll{\'a}r, and Ross Girshick.
\newblock Mask r-cnn.
\newblock In \emph{Proceedings of the IEEE international conference on computer vision}, pages 2961--2969, 2017.

\bibitem[Hendrycks and Dietterich(2019)]{hendrycks2019benchmarking}
Dan Hendrycks and Thomas Dietterich.
\newblock Benchmarking neural network robustness to common corruptions and perturbations.
\newblock In \emph{Proceedings of the International Conference on Learning Representations (ICLR)}, 2019.

\bibitem[Huang et~al.(2017)Huang, Liu, Van Der~Maaten, and Weinberger]{huang2017densely}
Gao Huang, Zhuang Liu, Laurens Van Der~Maaten, and Kilian~Q Weinberger.
\newblock Densely connected convolutional networks.
\newblock In \emph{Proceedings of the IEEE/CVF Conference on Computer Vision and Pattern Recognition (CVPR)}, 2017.

\bibitem[Jordahn and Olmos(2024)]{jordahn2024decoupling}
Mikkel Jordahn and Pablo~M. Olmos.
\newblock Decoupling feature extraction and classification layers for calibrated neural networks.
\newblock In \emph{Proceedings of the 41st International Conference on Machine Learning (ICML)}, 2024.

\bibitem[Kothapalli et~al.(2023)Kothapalli, Tirer, and Bruna]{kothapalli2023neural}
Vignesh Kothapalli, Tom Tirer, and Joan Bruna.
\newblock A neural collapse perspective on feature evolution in graph neural networks.
\newblock In \emph{Proceedings of the Advances in Neural Information Processing Systems (NeurIPS)}, 2023.

\bibitem[Krizhevsky et~al.(2009)Krizhevsky, Hinton, et~al.]{krizhevsky2009learning}
Alex Krizhevsky, Geoffrey Hinton, et~al.
\newblock Learning multiple layers of features from tiny images.
\newblock Technical report, University of Toronto, 2009.

\bibitem[Kumar et~al.(2018)Kumar, Sarawagi, and Jain]{kumar2018trainable}
Aviral Kumar, Sunita Sarawagi, and Ujjwal Jain.
\newblock Trainable calibration measures for neural networks from kernel mean embeddings.
\newblock In \emph{Proceedings of the 35th International Conference on Machine Learning (ICML)}, 2018.

\bibitem[Le and Yang(2015)]{le2015tiny}
Ya Le and Xuan Yang.
\newblock Tiny imagenet visual recognition challenge.
\newblock \emph{CS 231N}, 7\penalty0 (7):\penalty0 3, 2015.

\bibitem[Li et~al.(2023)Li, Liu, Zhou, Lu, Fernandez-Granda, Zhu, and Qu]{li2022understanding}
Xiao Li, Sheng Liu, Jinxin Zhou, Xinyu Lu, Carlos Fernandez-Granda, Zhihui Zhu, and Qing Qu.
\newblock Understanding and improving transfer learning of deep models via neural collapse.
\newblock \emph{Transactions on Machine Learning Research (TMLR)}, 2023.

\bibitem[Lin et~al.(2017)Lin, Goyal, Girshick, He, and Dollár]{lin2017focal}
Tsung-Yi Lin, Priya Goyal, Ross Girshick, Kaiming He, and Piotr Dollár.
\newblock Focal loss for dense object detection.
\newblock In \emph{Proceedings of the IEEE International Conference on Computer Vision (ICCV)}, 2017.

\bibitem[Liu et~al.(2022)Liu, Ben~Ayed, Galdran, and Dolz]{liu2022devil}
Bingyuan Liu, Ismail Ben~Ayed, Adrian Galdran, and Jose Dolz.
\newblock The devil is in the margin: Margin-based label smoothing for network calibration.
\newblock In \emph{Proceedings of the IEEE/CVF Conference on Computer Vision and Pattern Recognition (CVPR)}, 2022.

\bibitem[Liu et~al.(2023)Liu, Yu, Weller, and Sch{\"o}lkopf]{liu2023generalizing}
W. Liu, L. Yu, A. Weller, and B. Sch{\"o}lkopf.
\newblock Generalizing and decoupling neural collapse via hyperspherical uniformity gap.
\newblock In \emph{Proceedings of the International Conference on Learning Representations (ICLR)}, 2023.

\bibitem[Mukhoti et~al.(2020)Mukhoti, Kulharia, Sanyal, Golodetz, Torr, and Dokania]{mukhoti2020calibrating}
Jishnu Mukhoti, Viveka Kulharia, Amartya Sanyal, Stuart Golodetz, Philip Torr, and Puneet Dokania.
\newblock Calibrating deep neural networks using focal loss.
\newblock In \emph{Proceedings of the 34th Conference on Neural Information Processing Systems (NeurIPS)}, 2020.

\bibitem[Müller et~al.(2019)Müller, Kornblith, and Hinton]{muller2019when}
Rafael Müller, Simon Kornblith, and Geoffrey~E Hinton.
\newblock When does label smoothing help?
\newblock In \emph{Proceedings of the 33rd Conference on Neural Information Processing Systems (NeurIPS)}, 2019.

\bibitem[Netzer et~al.(2011)Netzer, Wang, Coates, Bissacco, Wu, and Ng]{netzer2011reading}
Yuval Netzer, Tao Wang, Adam Coates, Alessandro Bissacco, Bo Wu, and Andrew~Y Ng.
\newblock Reading digits in natural images with unsupervised feature learning.
\newblock In \emph{Proceedings of the 2011 International Conference on Machine Learning (ICML)}, 2011.

\bibitem[Nguyen and O'Connor(2015)]{nguyen2015posterior}
Khanh Nguyen and Brendan O'Connor.
\newblock Posterior calibration and exploratory analysis for natural language processing models.
\newblock \emph{arXiv preprint arXiv:1508.05154}, 2015.

\bibitem[Ovadia et~al.(2019)Ovadia, Fertig, Ren, Nado, Sculley, Nowozin, Dillon, Lakshminarayanan, and Snoek]{ovadia2019can}
Yaniv Ovadia, Emily Fertig, Jie Ren, Zachary Nado, David Sculley, Sebastian Nowozin, Joshua Dillon, Balaji Lakshminarayanan, and Jasper Snoek.
\newblock Can you trust your model's uncertainty? evaluating predictive uncertainty under dataset shift.
\newblock In \emph{Advances in neural information processing systems}, 2019.

\bibitem[Papyan et~al.(2020)Papyan, Han, and Donoho]{papyan2020prevalence}
Vardan Papyan, X.~Y. Han, and David~L. Donoho.
\newblock Prevalence of neural collapse during the terminal phase of deep learning training.
\newblock \emph{Proceedings of the National Academy of Sciences}, 117:\penalty0 24652--24663, 2020.

\bibitem[Park et~al.(2023)Park, Noh, Oh, Baek, and Ham]{park2023acls}
H. Park, J. Noh, Y. Oh, D. Baek, and B. Ham.
\newblock Acls: Adaptive and conditional label smoothing for network calibration.
\newblock In \emph{Proceedings of the IEEE/CVF International Conference on Computer Vision (ICCV)}, 2023.

\bibitem[Peifeng et~al.(2023)Peifeng, Xu, Wen, Yang, Shao, and Huang]{peifeng2023feature}
G. Peifeng, Q. Xu, P. Wen, Z. Yang, H. Shao, and Q. Huang.
\newblock Feature directions matters: Long-tailed learning via rotated balanced representation.
\newblock In \emph{Proceedings of the International Conference on Machine Learning (ICML)}, 2023.

\bibitem[Pinto et~al.(2022)Pinto, Yang, Lim, Torr, and Dokania]{pinto2022using}
Francesco Pinto, Harry Yang, Ser-Nam Lim, Philip Torr, and Puneet~K Dokania.
\newblock Using mixup as a regularizer can surprisingly improve accuracy \& out-of-distribution robustness.
\newblock In \emph{Proceedings of the 36th Conference on Neural Information Processing Systems (NeurIPS)}, 2022.

\bibitem[Redmon(2018)]{redmon2018yolov3}
Joseph Redmon.
\newblock Yolov3: An incremental improvement.
\newblock \emph{arXiv preprint arXiv:1804.02767}, 2018.

\bibitem[Shen et~al.(2019)Shen, Margolies, Rothstein, Fluder, McBride, and Sieh]{shen2019deep}
Li Shen, Laurie~R Margolies, Joseph~H Rothstein, Eugene Fluder, Russell McBride, and Weiva Sieh.
\newblock Deep learning to improve breast cancer detection on screening mammography.
\newblock \emph{Scientific reports}, 9\penalty0 (1):\penalty0 12495, 2019.

\bibitem[Szegedy et~al.(2016)Szegedy, Vanhoucke, Ioffe, Shlens, and Wojna]{szegedy2016rethinking}
Christian Szegedy, Vincent Vanhoucke, Sergey Ioffe, Jon Shlens, and Zbigniew Wojna.
\newblock Rethinking the inception architecture for computer vision.
\newblock In \emph{Proceedings of the IEEE Conference on Computer Vision and Pattern Recognition (CVPR)}, 2016.

\bibitem[Tao et~al.(2023{\natexlab{a}})Tao, Dong, Liu, Sun, and Xu]{tao2023calibrating}
Linwei Tao, Minjing Dong, Daochang Liu, Changming Sun, and Chang Xu.
\newblock Calibrating a deep neural network with its predecessors.
\newblock In \emph{Proceedings of the International Joint Conference on Artificial Intelligence (IJCAI)}, 2023{\natexlab{a}}.

\bibitem[Tao et~al.(2023{\natexlab{b}})Tao, Dong, and Xu]{tao2023dual}
Linwei Tao, Minjing Dong, and Chang Xu.
\newblock Dual focal loss for calibration.
\newblock In \emph{Proceedings of the 40th International Conference on Machine Learning (ICML)}, pages 33833--33849, 2023{\natexlab{b}}.

\bibitem[Thrampoulidis et~al.(2022)Thrampoulidis, Kini, Vakilian, and Behnia]{thrampoulidis2022imbalance}
Christos Thrampoulidis, Ganesh~R Kini, Vala Vakilian, and Tina Behnia.
\newblock Imbalance trouble: Revisiting neural-collapse geometry.
\newblock In \emph{Advances in Neural Information Processing Systems (NeurIPS)}, pages 27225--27238, 2022.

\bibitem[Thulasidasan et~al.(2019)Thulasidasan, Chennupati, Bilmes, Bhattacharya, and Michalak]{thulasidasan2019mixup}
Sunil Thulasidasan, Gopinath Chennupati, Jeff~A Bilmes, Tanmoy Bhattacharya, and Sarah Michalak.
\newblock On mixup training: Improved calibration and predictive uncertainty for deep neural networks.
\newblock In \emph{Proceedings of the 33rd Conference on Neural Information Processing Systems (NeurIPS)}, 2019.

\bibitem[Tirer and Bruna(2022)]{tirer2022extended}
Tom Tirer and Joan Bruna.
\newblock Extended unconstrained features model for exploring deep neural collapse.
\newblock In \emph{Proceedings of the 39th International Conference on Machine Learning (ICML)}, 2022.

\bibitem[Vaswani(2017)]{vaswani2017attention}
A Vaswani.
\newblock Attention is all you need.
\newblock \emph{Advances in Neural Information Processing Systems}, 2017.

\bibitem[Wang and Zhang(2024)]{wang2024calibration}
Deng-Bao Wang and Min-Ling Zhang.
\newblock Calibration bottleneck: Over-compressed representations are less calibratable.
\newblock In \emph{Proceedings of the 41st International Conference on Machine Learning (ICML)}, 2024.

\bibitem[Wang et~al.(2023)Wang, Li, Zhao, Heng, and Zhang]{wang2023pitfall}
Deng-Bao Wang, Lanqing Li, Peilin Zhao, Pheng-Ann Heng, and Min-Ling Zhang.
\newblock On the pitfall of mixup for uncertainty calibration.
\newblock In \emph{Proceedings of the IEEE/CVF Conference on Computer Vision and Pattern Recognition (CVPR)}, pages 7609--7618, 2023.

\bibitem[Xie et~al.(2023)Xie, Yang, Cai, and He]{xie2023neural}
Liang Xie, Yibo Yang, Deng Cai, and Xiaofei He.
\newblock Neural collapse inspired attraction-repulsion-balanced loss for imbalanced learning.
\newblock \emph{Neurocomputing}, 527:\penalty0 60--70, 2023.

\bibitem[Xu and Liu(2023)]{xu2023quantifying}
J. Xu and H. Liu.
\newblock Quantifying the variability collapse of neural networks.
\newblock In \emph{Proceedings of the International Conference on Machine Learning (ICML)}, 2023.

\bibitem[Yang et~al.(2022)Yang, Chen, Li, Xie, Lin, and Tao]{yang2022inducing}
Yibo Yang, Shixiang Chen, Xiangtai Li, Liang Xie, Zhouchen Lin, and Dacheng Tao.
\newblock Inducing neural collapse in imbalanced learning: Do we really need a learnable classifier at the end of deep neural network?
\newblock In \emph{Proceedings of the Advances in Neural Information Processing Systems (NeurIPS)}, 2022.

\bibitem[Yang et~al.(2023{\natexlab{a}})Yang, Steinhardt, and Hu]{yang2023are}
Y. Yang, J. Steinhardt, and W. Hu.
\newblock Are neurons actually collapsed? on the fine-grained structure in neural representations.
\newblock In \emph{Proceedings of the International Conference on Machine Learning (ICML)}, 2023{\natexlab{a}}.

\bibitem[Yang et~al.(2023{\natexlab{b}})Yang, Yuan, Li, Lin, Torr, and Tao]{yang2023neural}
Yibo Yang, Haobo Yuan, Xiangtai Li, Zhouchen Lin, Philip Torr, and Dacheng Tao.
\newblock Neural collapse inspired feature-classifier alignment for few-shot class-incremental learning.
\newblock In \emph{Proceedings of the International Conference on Learning Representations (ICLR)}, 2023{\natexlab{b}}.

\bibitem[Zagoruyko and Komodakis(2016)]{zagoruyko2016wide}
Sergey Zagoruyko and Nikos Komodakis.
\newblock Wide residual networks.
\newblock In \emph{Proceedings of the British Machine Vision Conference (BMVC)}, page~87, York, UK, 2016. BMVA Press.

\bibitem[Zhang et~al.(2018)Zhang, Cisse, Dauphin, and Lopez-Paz]{zhang2018mixup}
Hongyi Zhang, Moustapha Cisse, Yann~N. Dauphin, and David Lopez-Paz.
\newblock Mixup: Beyond empirical risk minimization.
\newblock In \emph{Proceedings of the International Conference on Learning Representations (ICLR)}, 2018.

\end{thebibliography}
}


\clearpage
\setcounter{page}{1}
\maketitlesupplementary

\begin{center}
\Large
\textbf{\thetitle}\\
\vspace{0.5em}Supplementary Material \\
\vspace{1.0em}
\end{center}
    
\section{The proof of Theorem 1}
\label{proof1}

As shown in \Cref{Simplex ETF}, within the general Simplex ETF framework, in addition to \( \beta \), the term \( \sqrt{\frac{K}{K-1}} \) in the matrix \( \mathbf{M} \) serves as a scaling factor, ensuring that each column vector has a fixed length. This scaling ensures that each column vector in \( \mathbf{M} \) has an equal norm, which is crucial for maintaining the isometric property required by a Simplex ETF. The scaling factor \( \beta \sqrt{\frac{K}{K-1}} \), plays a crucial role in shaping the model's probability distribution, output confidence, and ultimately its classification performance. A detailed analysis of this factor provides a deeper understanding of its adjustment mechanism and core functionality in classification models.

\paragraph{Class probability expression in softmax function.}

The softmax function is used to map the raw scores to class probabilities. The probability \( p_{i,k} \) for class \( k \) is given by:

\begin{equation}
p_{i,k} = \frac{\exp\left( \beta \sqrt{\frac{K}{K-1}} \bm{z}_i \left[ \mathbf{U} \left( \mathbf{I} - \frac{1}{K} \mathbf{1}_K \mathbf{1}_K^\mathrm{T} \right) \right]_k \right)}{\sum_{j=1}^{K} \exp\left( \beta \sqrt{\frac{K}{K-1}} \bm{z}_i \left[ \mathbf{U} \left( \mathbf{I} - \frac{1}{K} \mathbf{1}_K \mathbf{1}_K^\mathrm{T} \right) \right]_j \right)}.
\end{equation}

To simplify the notation, we define \( \sigma_i = \bm{z}_i \hat{\bm{m}}_i \), where \( \hat{\bm{m}}_i \) is the \( i \)-th column vector of the matrix \( \mathbf{U} \left( \mathbf{I} - \frac{1}{K} \mathbf{1}_K \mathbf{1}_K^\top \right) \). The class probability expression then simplifies to:

\begin{equation}
p_{i,k} = \frac{\exp\left( \beta \sqrt{\frac{K}{K-1}} \sigma_i \right)}{\sum_{j=1}^{K} \exp\left( \beta \sqrt{\frac{K}{K-1}} \sigma_j \right)}.
\end{equation}

Further neglecting the normalization constant \( \sum_{j=1}^{K} \exp\left( \beta \sqrt{\frac{K}{K-1}} \sigma_j \right) \), the predicted confidence \( \hat{p}_i \) is defined as \( \hat{p}_i = \max_k p_{i,k} \), which can be approximated as:

\begin{equation}
\hat{p}_i \propto \exp\left( \beta \sqrt{\frac{K}{K-1}} \sigma_i \right).
\end{equation}

This equation clearly demonstrates the critical role of the scaling factor \( \beta \sqrt{\frac{K}{K-1}} \) in the Softmax function. The magnitude of the scaling factor directly determines the concentration of the class probability distribution: a larger \( \beta \sqrt{\frac{K}{K-1}} \) significantly increases the probability of the higher-scoring class, sharpening the confidence for that class; a smaller value leads to a more uniform probability distribution, increasing the uncertainty of the output of the model.

\paragraph{Impact of scaling factor: Changes in class probability ratios.}

Further analysis of the relative probability ratio between the predicted class \( i \) and other classes \( j \) is formulated as:

\begin{equation}
\frac{\hat{p}_i}{p_j} = \exp\left( \beta \sqrt{\frac{K}{K-1}} (\sigma_i - \sigma_j) \right).
\end{equation}

This equation indicates that the relative relationship between the probabilities of classes is governed by the difference \( \sigma_i - \sigma_j \), which is scaled by the factor \( \beta \sqrt{\frac{K}{K-1}} \). In other words, the scaling factor adjusts the impact of the score difference on the final classification decision by amplifying the difference in \( \sigma_i - \sigma_j \), thereby strengthening the model's ability to discriminate between classes.

It is important to note that while the scaling factor does not alter the absolute ranking of class probabilities, it significantly adjusts the relative relationships between probabilities, making the model more sensitive to the geometric structure of the classes and their distinctions.

\paragraph{Limit behavior analysis: from random guessing to determined.}
To better understand the impact of the scaling factor, consider its behavior in the limiting cases:

1. \textbf{When \( \beta \sqrt{\frac{K}{K-1}} \to 0 \)}, the class probability distribution tends towards uniformity:
   \begin{equation}
   \lim_{\beta \sqrt{\frac{K}{K-1}} \to 0}p_{i,k} = \frac{1}{\sum_{j=1}^{K}1} = \frac{1}{K}.
   \end{equation}
   In this case, the probabilities for all classes become equal, and the model behaves as if making random guesses, unable to effectively distinguish between classes.

2. \textbf{When \( \beta \sqrt{\frac{K}{K-1}} \to \infty \)}, the Softmax function approaches the Argmax function, and the class probability distribution becomes nearly deterministic:
   \begin{equation}
   \lim_{\beta \sqrt{\frac{K}{K-1}} \to \infty}\hat{p}_i = \frac{1}{1+\sum_{j\neq i}^{K} \exp( \beta \sqrt{\frac{K}{K-1}} (\sigma_j - \sigma_i ))} = 1,
   \end{equation}
   where \( i \) is the class with the highest score. In this extreme case, the model’s output becomes nearly deterministic, always predicting the highest-scoring class.

\paragraph{Role of scaling factor: from adjustment to optimization.}
The scaling factor \( \beta \sqrt{\frac{K}{K-1}} \) in the model serves more than just an adjustment function for the class probability distribution. It amplifies the score differences \( \sigma_i - \sigma_j \), thereby enhancing the model’s ability to distinguish between classes. Specifically, by enlarging the differences in class scores, the scaling factor increases the probability of the higher-scoring classes, improving the model's discriminative ability. When the factor is smaller, the class probabilities become more uniform, leading to weaker classification performance.

Thus, the flexible adjustment of the scaling factor allows the model to appropriately calibrate the classification confidence across different tasks and data distributions. The optimization of this factor not baseline makes the Softmax function's output more flexible but also leverages the geometric structure of the classes to enhance confidence calibration.

\section{Implementation details}

\label{Implementation}
\paragraph{Experimental environment and configuration.}

All experiments are conducted on the Ubuntu 20.04.4 LTS operating system, Intel(R) Xeon(R) Gold 5220 CPU @ 2.20GHz with a single NVIDIA A40 48GB GPU, and 512GB of RAM. The framework is implemented with Python 3.8.19 and PyTorch 2.0.1. Other key packages include numpy 1.23.5, pandas 2.0.3, and scipy 1.10.1.

\paragraph{Datasets.}
We conduct model calibration experiments on various benchmark datasets, including CIFAR-10, CIFAR-100 \citep{krizhevsky2009learning}, SVHN \citep{netzer2011reading}, and Tiny-ImageNet \citep{le2015tiny} aiming to evaluate calibration techniques across a spectrum of data complexities and challenges. 
CIFAR-10 and CIFAR-100 consist of \(32 \times 32\) RGB images of 10 and 100 classes, respectively, with 50,000 training and 10,000 test samples. They are widely used as benchmarks for computer vision tasks, offering a controlled and well-understood testing environment for model calibration. SVHN comprises \(32 \times 32\) images of digits (0–9) extracted from real-world street view scenes, containing 73,257 training and 26,032 test images. Its inherent variability and real-world noise introduce unique challenges for calibration studies. Tiny-ImageNet, a subset of the larger ImageNet dataset, includes \(64 \times 64\) images spanning 200 classes, with 500 training, 50 validation, and 50 test samples per class. Compared to CIFAR datasets, Tiny-ImageNet demands greater generalization capabilities due to its higher resolution, broader category set, and limited training samples per class. Overall, these datasets provide a range of conditions to systematically evaluate calibration methods under varying levels of data complexity and difficulty.

To assess model robustness, we explore calibration performance under distributional shifts and out-of-distribution (OOD) scenarios. For distributional shifts, CIFAR-10 and CIFAR-100 are evaluated on their corrupted versions, CIFAR-10-C and CIFAR-100-C \citep{hendrycks2019benchmarking}, which feature 15 corruption types (e.g., noise, blur, brightness) at 5 severity levels. 
For OOD detection, we test models trained on each dataset against samples from other datasets, leveraging their distinct visual and semantic characteristics. For instance, models trained on SVHN are evaluated on CIFAR-10 and CIFAR-100, while those trained on CIFAR datasets are tested on each other and SVHN. This comprehensive evaluation framework highlights the interplay between calibration quality and model generalization.


\paragraph{Training details.}
Following \citep{jordahn2024decoupling}, we train a Wide Residual Network (WRN) 28-10 \cite{zagoruyko2016wide} across all datasets. Training experiments are conducted for 600 epochs using the Adam optimizer with a learning rate of \(10^{-4}\), applying early stopping based on validation loss. All validation experiments are derived from the training set by reserving 15\% of the samples for CIFAR-10 and SVHN, and 5\% for CIFAR-100. 
To further evaluate our method, we also train ResNet50 \citep{He_2016_CVPR} and ResNet101 \citep{He_2016_CVPR} backbones using the setting \citep{liu2022devil} for calibration performance. We employ the SGD optimizer with a momentum of 0.9 and an initial learning rate of \(0.1\). The learning rate is reduced by factor 10 at predefined intervals during training. 

\paragraph{Reproduced details.} We provide reproduced hyperparameter  details of baseline methods as follows: 
(a) LS \cite{szegedy2016rethinking}: Following \cite{mukhoti2020calibrating}, we report the results obtained with $\alpha = 0.05$. 
(b) FL \citep{mukhoti2020calibrating}: We train the models with a fixed regularization parameter $\gamma=3$. 
(c) Mixup \cite{thulasidasan2019mixup}: We use a hyperparameter $\alpha$ of 0.2, the best performing one in \cite{thulasidasan2019mixup}. 
(d) MIT \cite{wang2023pitfall}: We specifically use MIT-L, with hyperparameter $\alpha = 1.0$, as applied in the main experiments of \cite{wang2023pitfall}.
(e) FLSD \citep{mukhoti2020calibrating}: Following the schedule in \citep{mukhoti2020calibrating}, we use the parameter $\gamma=5$ for the samples whose output probability for the ground-truth class is within [0, 0.2), otherwise we use the parameter $\gamma=3$. 
(f) CPC \citep{Cao2019}: We set the weights of binary discrimination and binary exclusion losses as 0.1 and 1.0, respectively. 
(g) ACLS \cite{park2023acls}: We set $\lambda_{1} = 1.0$, $\lambda_{2} = 0.1$, margin $mgn = 10.0$, and $\alpha = 0.1$ for general experiments, but adjusted $mgn$ to 6.0 for CIFAR-10 to better suit its characteristics. 
(h) MMCE \cite{kumar2018trainable}: We report the results obtained with $\alpha = 0.5$ in CIFAR100 and $\alpha = 1.5$ in CIFAR10 and SVHN. 
(i) PLP \cite{wang2024calibration}: We set hyperparameter $\gamma$, which determines the starting epoch at which the top layers begin to be frozen, to 1.25, as mentioned in \cite{wang2024calibration}. 
(j) TST and VTST \cite{jordahn2024decoupling}: We set the number of neurons in the final hidden layer output to 128.

\paragraph{Evaluation protocols.}
\label{Evaluation protocols.}
We follow standard protocols \citep{guo2017calibration,nguyen2015posterior} and utilize the Expected Calibration Error (ECE), and Adaptive Expected Calibration Error (AECE) for evaluating network calibration performance. Expected Calibration Error (ECE) quantifies the expected $\mathbb{E}\left[\left|P(\hat{{y}}_{i}={y}_{i} | \hat{{p}}_{i})-\hat{p}_{i}\right|\right]$ of the absolute difference between predicted confidence and actual accuracy.
ECE is calculated by dividing the confidence range $(0, 1]$ into 15 equally spaced bins and averaging the absolute differences within each bin, weighted by bin size:
\begin{equation}
\label{ece}
\text{ECE} = \sum_{m=1}^{M}\frac{|B_m|}{n}\bigg|Acc(B_m) - Conf(B_m)\bigg|,
\end{equation}
where $|B_m|$ is the number of samples in bin $m$, $n$ is the total number of samples, $Acc(B_m)$ denotes the accuracy within bin $B_m$, and $Conf(B_m)$ represents the average confidence within the bin. Bins are uniformly spaced over the range $(0, 1]$, ensuring a consistent confidence interval width of $1/M$. Following standard practice, we set bins $M = 15$. The difference between $Acc$ and $Conf$ can indicate the calibration gap for model calibration. Adaptive Expected Calibration Error (AECE) enhances ECE by dynamically adjusting the bin boundaries through the confidence distribution of the samples.
In contrast to ECE with fixed bins, AECE employs an adaptive binning strategy where bin sizes are determined to evenly distribute samples across bins. This approach facilitates a more effective assessment of calibration performance, particularly for imbalanced or skewed confidence distributions.
Besides, Top-1 classification accuracy is reported for discriminative evaluation. 
For OOD detection, we also compare AUROC and FPR95, adhering to standard evaluation protocols \citep{pinto2022using}.

\begin{table}[htbp]
  \centering
  \resizebox{0.85\textwidth}{!}{
    \begin{tabular}{c|cccc|cccc|cccc}
    \toprule
          & \multicolumn{4}{c|}{\textbf{CIFAR-10}}  & \multicolumn{4}{c|}{\textbf{CIFAR-100}} & \multicolumn{4}{c}{\textbf{SVHN}} \\
\cmidrule{2-13}          & \multicolumn{2}{c}{baseline} & \multicolumn{2}{c|}{+Ours} & \multicolumn{2}{c}{baseline} & \multicolumn{2}{c|}{+BalCAL} & \multicolumn{2}{c}{baseline} & \multicolumn{2}{c}{+BalCAL} \\
\cmidrule{2-13}          & Acc↑  & \multicolumn{1}{c|}{ECE↓} & Acc↑  & ECE↓  & Acc↑  & \multicolumn{1}{c|}{ECE↓} & Acc↑  & \multicolumn{1}{c|}{ECE↓} & Acc↑  & \multicolumn{1}{c|}{ECE↓} & Acc↑  & ECE↓ \\
    \midrule
    ACLS\citep{park2023acls}  & 93.12 & 5.26  & 93.07$\textcolor{red}\downarrow$  & 2.10$\textcolor{green}\downarrow$ & 72.36  & 13.08  & 72.36$\textcolor{green}\uparrow$ & 2.07$\textcolor{green}\downarrow$ & 95.33  & 3.20  & 95.35$\textcolor{green}\uparrow$ & 1.47$\textcolor{green}\downarrow$ \\
    LS\citep{szegedy2016rethinking}    & 92.12  & 2.79  & 92.16$\textcolor{green}\uparrow$ & 2.88$\textcolor{red}\uparrow$  & 73.58  & 6.03  & 73.60$\textcolor{green}\uparrow$ & 3.86$\textcolor{green}\downarrow$ & 95.50& 3.77  & 95.52$\textcolor{red}\downarrow$ & 1.52$\textcolor{green}\downarrow$ \\
     FL\citep{mukhoti2020calibrating}    & 92.11 & 1.99  & 92.06$\textcolor{red}\downarrow$  & 1.32$\textcolor{green}\downarrow$ & 71.54  & 14.26  & 72.06$\textcolor{green}\uparrow$ & 1.16$\textcolor{green}\downarrow$ & 94.79 & 1.24  & 94.71$\textcolor{red}\downarrow$  & 1.70$\textcolor{red}\uparrow$  \\
    FLSD\citep{mukhoti2020calibrating}  & 92.42 & 1.72  & 92.35$\textcolor{red}\downarrow$  & 1.16$\textcolor{green}\downarrow$ & 71.19  & 14.20  & 71.59$\textcolor{green}\uparrow$ & 1.08$\textcolor{green}\downarrow$ & 94.60  & 1.03  & 94.68$\textcolor{green}\uparrow$ & 1.74$\textcolor{green}\downarrow$ \\
    CPC\citep{Cao2019}   & 91.56  & 6.28  & 91.71$\textcolor{green}\uparrow$ & 2.47$\textcolor{green}\downarrow$ & 73.33 & 12.53  & 73.31$\textcolor{red}\downarrow$  & 7.95$\textcolor{green}\downarrow$ & 94.73  & 1.61  & 94.83$\textcolor{green}\uparrow$ & 0.85$\textcolor{green}\downarrow$ \\
    MMCE\citep{kumar2018trainable}  & 90.04  & 3.60  & 90.24$\textcolor{green}\uparrow$ & 1.72$\textcolor{green}\downarrow$ & 68.77  & 17.18  & 70.28$\textcolor{green}\uparrow$ & 3.09$\textcolor{green}\downarrow$ & 93.99  & 2.30 & 94.60$\textcolor{green}\uparrow$ & 2.76$\textcolor{red}\uparrow$  \\
    Mixup\citep{thulasidasan2019mixup} & 94.71  & 2.56  & 94.72$\textcolor{green}\uparrow$ & 1.25$\textcolor{green}\downarrow$ & 77.05 & 4.51  & 76.40$\textcolor{red}\downarrow$  & 2.27$\textcolor{green}\downarrow$ & 94.82& 3.75  & 92.38$\textcolor{red}\downarrow$ & 1.46$\textcolor{green}\downarrow$ \\
    MIT\citep{wang2023pitfall}   & 94.27  & 2.18  & 94.72$\textcolor{green}\uparrow$ & 0.69$\textcolor{green}\downarrow$ & 73.38  & 12.31  & 75.05$\textcolor{green}\uparrow$ & 3.06$\textcolor{green}\downarrow$ & 93.69  & 0.78  & 94.22$\textcolor{green}\uparrow$ & 0.75$\textcolor{green}\downarrow$ \\
    \bottomrule
    \end{tabular}%
  }
    \caption{Results of combination ours with commonly baseline calibration methods. Green arrows indicate improved performance, while red arrows indicate decreased performance.}
  \label{tab:combination}
  
\end{table}

\section{More evaluation results}
\label{MoreResults}

\paragraph{Combination ours with baseline calibration methods.}
We combine BalCAL with common baseline calibration methods, which consist of loss-based(ACLS\citep{park2023acls}, LS\citep{szegedy2016rethinking}, FL\citep{mukhoti2020calibrating}, FLSD\citep{mukhoti2020calibrating}, CPC\citep{Cao2019}, MMCE\citep{kumar2018trainable}) and augment-based (Mixup\citep{thulasidasan2019mixup} and MIT\citep{wang2023pitfall}) methods in \Cref{tab:combination}. 
We can find that model performance presents improvements on most baseline methods, especially ECE. 
This indicates that our method has strong compatibility and the potential to enhance model performance.

\paragraph{Extended Architectures and Datasets}
We conducted extensive experiments on additional model architectures, including WRN-26-10 \citep{zagoruyko2016wide}, DenseNet-121 \citep{huang2017densely}, and ViT-B\_16 \citep{dosovitskiy2020image}, as well as on larger datasets such as ImageNet \citep{deng2009imagenet}. We also incorporated new comparative methods \citep{tao2023dual,tao2023calibrating} to further validate our approach. The results are presented in \cref{tab:architectures} and \cref{tab:vit}, where the experimental setup for \cref{tab:architectures} follows \citep{mukhoti2020calibrating}, and the setup for \cref{tab:vit} adheres to \citep{dosovitskiy2020image}. Our method consistently demonstrates superior performance across all evaluated scenarios.

\begin{table}[H]
  \centering
  \resizebox{0.5\linewidth}{!}{
    \begin{tabular}{c|cccc|cccc}
    \toprule
    \multirow{3}[6]{*}{\textbf{Method}} & \multicolumn{4}{c|}{\textbf{WRN-26-10}} & \multicolumn{4}{c}{\textbf{DenseNet-121}} \\
\cmidrule{2-9}          & \multicolumn{2}{c}{CIFAR-10} & \multicolumn{2}{c|}{CIFAR-100} & \multicolumn{2}{c}{CIFAR-10} & \multicolumn{2}{c}{CIFAR-100} \\
\cmidrule{2-9}          & Acc↑ & ECE↓ & Acc↑ & ECE↓ & Acc↑ & ECE↓ & Acc↑ & ECE↓ \\
    \midrule
    Vanilla & 96.03 & 3.37  & 79.82 & 10.98 & 95.05 & 4.02  & 77.73 & 12.07 \\
    \textbf{DualFocal$^*$} & 96.04 & 0.81  & 80.09 & 1.79  & 94.57 & 0.57  & 77.60  & 1.81 \\
    \textbf{PCS$^*$} & -     & 0.99  & -     & 1.92  & -     & 0.78  & -     & 2.75 \\
    \rowcolor{gray!20} 
    Ours & \textbf{96.24} & \textbf{0.79} & \textbf{80.27} & \textbf{1.51} & \textbf{95.35} & \textbf{0.53} & \textbf{77.81} & \textbf{1.77} \\
    \bottomrule
    \end{tabular}%
    }
    \caption{Calibration Performance Across Diverse Model Architectures. $^*$ denotes paper results.}
  \label{tab:architectures}
\end{table}
\vspace{-1.2em}

\begin{table}[H]
  \centering
   \resizebox{0.45\linewidth}{!}{
    \begin{tabular}{c|ccc|ccc}
    \toprule
    \multirow{2}[4]{*}{\textbf{Methods}} & \multicolumn{3}{c|}{CIFAR-100} & \multicolumn{3}{c}{\textbf{ImageNet}} \\
\cmidrule{2-7}          & Acc↑ & ECE↓ & \textbf{AECE↓} & Acc↑ & ECE↓ & \textbf{AECE↓} \\
    \midrule
    Vanilla & 92.67  & 1.75  & 1.45  & 81.85  & 5.74  & 5.72  \\
    \textbf{MMCE\citep{kumar2018trainable}} & \textbf{92.76} & 1.27  & 1.08  & 81.68  & 4.91  & 4.86  \\
    \textbf{DualFocal} & 92.73  & 1.25  & 1.05  & 81.94  & 1.69  & 1.61  \\
    \textbf{TST\citep{jordahn2024decoupling}} & 92.10  & 1.09  & 1.27  & 81.04  & 4.38  & 4.31  \\
    \textbf{VTST\citep{jordahn2024decoupling}} & 92.67  & 1.49  & 1.34  & 80.00  & 2.01  & 1.91  \\
    \rowcolor{gray!20}
    Ours & 92.69  & \textbf{0.96} & \textbf{0.97} & \textbf{82.32} & \textbf{1.48} & \textbf{1.49} \\
    \bottomrule
    \end{tabular}
  }  
  \caption{Calibration Analysis on ViT-B\_16.}
  \label{tab:vit}
\end{table}

\paragraph{{Uncertainty evaluation in OOD detection.}}
In \Cref{tab:ood22222}, we examine model confidence and entropy for OOD samples. Ideally, OOD samples should present low confidence and high entropy, reflecting the model's caution with unknown data. Our method can align with this trend and achieve the best performance with lower confidence and higher entropy, indicating significant uncertainty regarding unfamiliar samples. Such uncertainty enhances the model's self-correction ability, improving adaptability in OOD detection and ensuring reliable performance across diverse samples.

\begin{table}[htbp]
  \centering
    \resizebox{0.86\textwidth}{!}{
    \begin{tabular}{c|cccc|cccc|cccc}
    \toprule
          \multirow{3}[2]{*}{\textbf{Methods}} & \multicolumn{4}{c|}{\textbf{CIFAR-10}}  & \multicolumn{4}{c|}{\textbf{CIFAR-100}} & \multicolumn{4}{c}{\textbf{SVHN}} \\
\cmidrule{2-13}          & \multicolumn{2}{c}{SVHN} & \multicolumn{2}{c|}{CIFAR-100} & \multicolumn{2}{c}{SVHN} & \multicolumn{2}{c|}{CIFAR-10} & \multicolumn{2}{c}{CIFAR-10} & \multicolumn{2}{c}{CIFAR-100} \\
\cmidrule{2-13}          & Conf↓ & Entropy↑ & Conf↓ & Entropy↑ & Conf↓ & Entropy↑ & Conf↓ & Entropy↑ & Conf↓ & Entropy↑ & Conf↓ & Entropy↑ \\
    \midrule
    Vanilla & 86.35  & 34.96  & 89.69  & 27.02  & 86.17  & 40.02  & 83.24  & 47.72  & 81.65  & 51.08  & 81.38  & 51.41  \\
    ACLS\citep{park2023acls}  & 90.55  & 26.74  & 90.73  & 26.56  & 68.33  & 134.92  & 67.73  & 136.49  & 77.95  & 66.44  & 79.18  & 62.84  \\
    LS\citep{szegedy2016rethinking}    & 85.76  & 53.63  & 85.33  & 55.88  & 52.89  & 238.43  & 55.55  & 226.29  & 65.93  & 116.51  & 67.47  & 112.34  \\
    mixup\citep{thulasidasan2019mixup} & 82.04  & 56.27  & 81.51  & 61.30  & 64.21  & 149.04  & 59.25  & 175.28  & 60.38  & 121.14  & 62.16  & 116.57  \\
    MIT\citep{wang2023pitfall}   & 71.20  & 77.50  & 79.03  & 56.81  & 59.67  & 133.32  & 67.18  & 106.62  & 79.03  & 56.37  & 77.40  & 60.68  \\
    FL\citep{mukhoti2020calibrating}    & 75.52  & 63.65  & 75.07  & 65.44  & 69.43  & 96.71  & 70.82  & 89.72  & 62.97  & 106.21  & 63.79  & 104.49  \\
    FLSD\citep{mukhoti2020calibrating}  & 73.15  & 69.53  & 75.42  & 64.58  & 69.10  & 101.06  & 70.10  & 91.70  & 66.39  & 95.83  & 66.45  & 95.61  \\
    \midrule
    CPC\citep{Cao2019}   & 89.64  & 37.11  & 86.97  & 45.70  & 66.52  & 171.02  & 60.79  & 200.46  & 57.38  & 140.76  & 57.67  & 140.22  \\
    PLP\citep{wang2024calibration}   & 87.55  & 32.94  & 86.85  & 35.59  & 70.25  & 110.65  & 68.29  & 113.99  & 81.84  & 50.55  & 82.32  & 49.34  \\
    MMCE\citep{kumar2018trainable}  & 66.51  & 97.36  & 71.26  & 84.60  & 68.71  & 102.19  & 71.71  & 91.51  & 36.41  & 185.99  & 36.84  & 184.77  \\
    TST\citep{jordahn2024decoupling}   & 68.51  & 95.73  & 72.08  & 84.35  & 60.95  & 144.44  & 58.13  & 149.60  & 61.80  & 118.68  & 61.93  & 118.30  \\
    VTST\citep{jordahn2024decoupling}  & 76.34  & 71.87  & 77.12  & 70.79  & 54.50  & 191.72  & 54.78  & 184.83  & 70.71  & 87.10  & 70.99  & 86.85  \\
    \rowcolor{gray!20} 
    \textbf{Ours} & \textbf{64.74}  & \textbf{120.08} & \textbf{68.79}  & \textbf{105.86} & \textbf{36.66}  & \textbf{305.63} & \textbf{45.07}  & \textbf{257.74} & \textbf{60.27} & \textbf{130.03} & \textbf{60.97} & \textbf{127.84} \\
    \bottomrule
    \end{tabular}%
}
  \caption{Confidence and entropy analysis for OOD samples.}
  \label{tab:ood22222}
\end{table}%

\paragraph{Underconfidence problem with mixup}

Mixup is a data augmentation method that improves accuracy by taking convex combinations of pairs of examples and labels. Given a sample $(x_i, y_i)$, Mixup generates a new sample by mixing it and another sample $(x_j, y_j)$ as follows:
\begin{equation}
\tilde{x} = \lambda x_i + (1 - \lambda) x_j, \quad \tilde{y} = \lambda y_i + (1 - \lambda) y_j,
\end{equation}
where $(x_i, y_i)$ and $(x_j, y_j)$ are randomly sampled from the training set $D$, and $\lambda \in [0, 1]$ is a mixing coefficient sampled from the Beta distribution $\text{Beta}(\alpha, \alpha)$. This process applies the same mixing coefficient $\lambda$ to both the input samples and their corresponding labels. The hyperparameter $\alpha$ controls the interpolation strength between the input pairs and their labels.
\\
\begin{figure}[H]
  \centering

  \begin{subfigure}{0.33\linewidth}
    \includegraphics[width=1.0\linewidth]{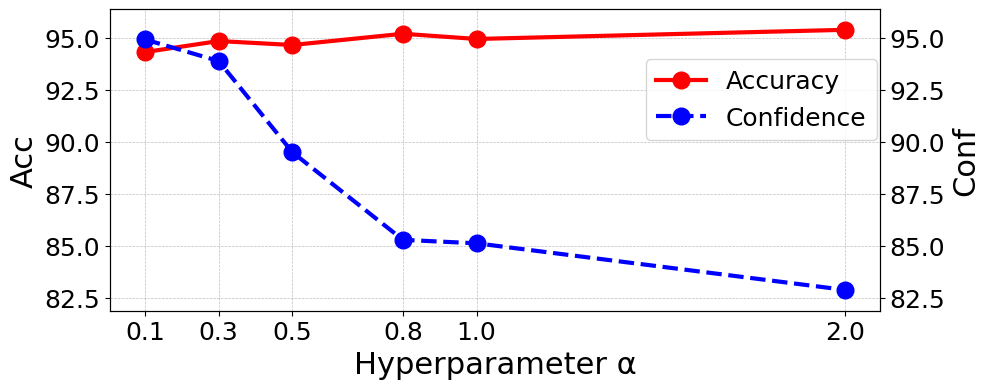}
    \caption{CIFAR-10}
  \end{subfigure}
  \hfill
  \begin{subfigure}{0.33\linewidth}
    \includegraphics[width=1.0\linewidth]{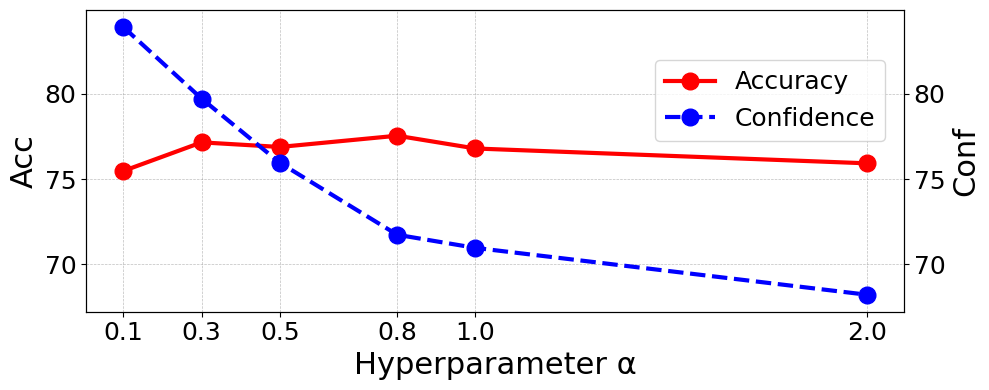}
    \caption{CIFAR-100}
  \end{subfigure}
  \hfill
  \begin{subfigure}{0.33\linewidth}
    \includegraphics[width=1.0\linewidth]{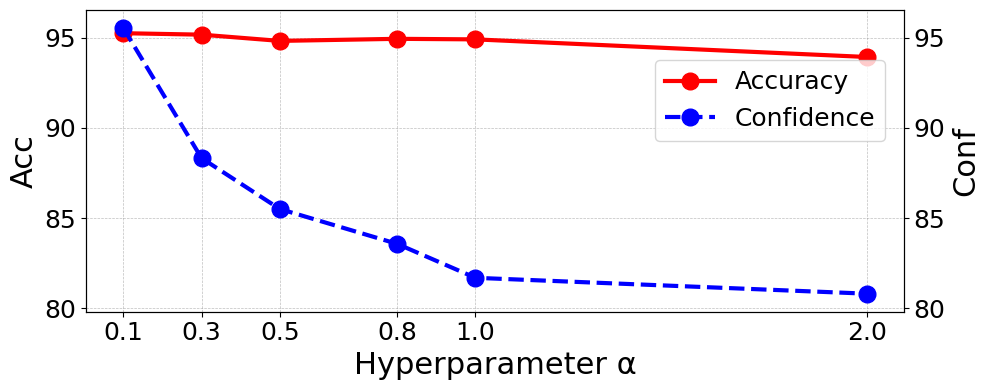}
    \caption{SVHN}
  \end{subfigure}

  \caption{Effect of $\alpha$ on model confidence and underconfidence.}
  \label{fig:mixup2}
\end{figure}

Recent work (MIT) \cite{wang2023pitfall} has indicated that Mixup improves model calibration performance but can lead to underconfidence problems. 
As illustrated in \cref{fig:mixup2}, we can find that when $\alpha$ becomes large, model confidence rapidly decreases and exhibits the underconfidence phenomenon. 
To investigate whether our approach effectively mitigates the underconfidence problem with Mixup, we evaluate the accuracy and ECE results of Mixup, MIT, and Mixup+Ours with various settings in \cref{tab:mixup}.
Mixup+Ours method significantly surpasses Mixup and MIT under different ratios $\alpha$ across various datasets.


\begin{table}[htbp]
  \centering
  \resizebox{0.53\textwidth}{!}{
    \begin{tabular}{cc|cc|cc|cc}
    \toprule
    \multirow{2}[2]{*}{\textbf{Dataset}} & \multirow{2}[2]{*}{$\boldsymbol{\alpha}$} & \multicolumn{2}{c|}{Mixup} & \multicolumn{2}{c|}{MIT} & \multicolumn{2}{c}{\textbf{Mixup+Ours}} \\
          &       & ACC↑  & ECE↓  & ACC↑  & ECE↓  & ACC↑  & ECE↓ \\
    \midrule
    \multirow{6}[2]{*}{\textbf{CIFAR-10}} & 0.1   & 94.30  & 2.39  & 93.67  & 3.62  & \textbf{94.72} & \textbf{1.25} \\
          & 0.3   & 94.83  & 3.38  & 94.40  & 2.62  & \textbf{95.28} & \textbf{1.13} \\
          & 0.5   & 94.64  & 6.00  & 93.29  & 2.97  & \textbf{94.78} & \textbf{1.39} \\
          & 0.8   & \textbf{95.18} & 10.22  & 93.60  & 2.68  & 95.05  & \textbf{1.71} \\
          & 1.0   & 94.93  & 9.93  & 94.27  & 2.18  & \textbf{95.06} & \textbf{1.21} \\
          & 2.0   & \textbf{95.37} & 12.50  & 93.10  & 2.61  & 95.13  & \textbf{1.37} \\
    \midrule
    \multirow{6}[2]{*}{\textbf{CIFAR-100}} & 0.1   & 75.48  & 8.48  & 73.48  & 14.18  & \textbf{76.40} & \textbf{2.27} \\
          & 0.3   & 77.16  & 2.64  & 74.23  & 12.41  & \textbf{78.22} & \textbf{1.89} \\
          & 0.5   & 76.89  & 3.97  & 74.55  & 11.11  & \textbf{78.15} & \textbf{2.47} \\
          & 0.8   & 77.55  & 6.12  & 73.25  & 12.00  & \textbf{77.88} & \textbf{2.93} \\
          & 1.0   & 76.80  & 6.43  & 73.38  & 12.31  & \textbf{78.06} & \textbf{2.16} \\
          & 2.0   & 75.93  & 8.64  & 72.16  & 11.44  & \textbf{76.68} & \textbf{1.67} \\
    \midrule
    \multirow{6}[2]{*}{\textbf{SVHN}} & 0.1   & 95.25  & 1.65  & 95.02  & 1.76  & \textbf{95.47} & \textbf{1.46} \\
          & 0.3   & 95.17  & 6.92  & 94.21  & \textbf{0.55} & \textbf{95.37} & 1.13  \\
          & 0.5   & 94.82  & 9.36  & 94.10  & 1.14  & \textbf{95.39} & \textbf{1.13} \\
          & 0.8   & 94.94  & 11.46  & 93.91  & \textbf{0.83} & \textbf{95.80} & 1.01  \\
          & 1.0   & 94.91  & 13.22  & 93.69  & \textbf{0.78} & \textbf{95.84} & 1.30  \\
          & 2.0   & 93.93  & 13.13  & 93.11  & \textbf{1.43} & \textbf{95.65} & 3.16  \\
    \bottomrule
    \end{tabular}%
   }
    \caption{Comparison Performance across datasets under different ratios $\alpha$.}
  \label{tab:mixup}
\end{table}

\paragraph{Computational cost} 
We rigorously evaluate the training efficiency by comparing the per-epoch time costs between our method and vanilla implementations across multiple architectures and datasets. As systematically documented in \cref{tab:cost}, our approach improves calibration performance at an acceptable cost.
\begin{table}[htbp]
  \centering
  \resizebox{0.45\linewidth}{!}{
    \begin{tabular}{c|cc|cc}
    \toprule
    \multirow{2}[4]{*}{} & \multicolumn{2}{c|}{\textbf{WRN-26-10}} & \multicolumn{2}{c}{\textbf{DenseNet-121}} \\
\cmidrule{2-5}          & CIFAR-10 & CIFAR-100 & CIFAR-10 & CIFAR-100 \\
    \midrule
    Vanilla & 63.06  & 63.84  & 54.01  & 54.43  \\
    Ours & 65.73  & 67.64  & 55.91  & 57.79  \\
    \bottomrule
    \end{tabular}%
    }
  \caption{Computational cost (s) per training epoch.}
  \label{tab:cost}%
\end{table}%

\paragraph{Dynamic and Fixed $\gamma$:}  
In this paper, we employ a dynamic $\gamma$, derived from \cref{equation:gamma}. The parameter $\gamma$ in \cref{equation:loss_total} and \cref{equation:p_fused} balances the learnable classifier and the fixed ETF classifier. Our ablation study in \cref{tab:gamma} indicates that $\gamma$ in $L_{\text{total}}$ is relatively insensitive, and a fixed $\gamma$ (e.g., 0.5) achieves competitive performance. However, using a dynamic $\gamma$ in $L_{\text{total}}$ provides slight improvements without additional computational cost.  
In contrast, for $\bm{p}^{fused}$, $\gamma$ is more sensitive and dataset-dependent, making a dynamic $\gamma$ essential for performance enhancement. To ensure consistency, we adopt the same dynamic $\gamma$ for both $L_{\text{total}}$ and $\bm{p}^{fused}$, although its impact is more pronounced in $\bm{p}^{fused}$.

\begin{table}[htbp]
  \centering
  \resizebox{0.5\linewidth}{!}{
    \begin{tabular}{c|c|cccccc}
    \toprule
    $\gamma$ of & $\gamma$ of & \multicolumn{2}{c}{CIFAR-10} & \multicolumn{2}{c}{CIFAR-100} & \multicolumn{2}{c}{\textbf{SVHN}} \\
\cmidrule{3-8}    \(L_{\text{total}}\) & \(\bm{p}^{fused}\) & ACC↑  & ECE↓  & ACC↑  & ECE↓  & ACC↑  & ECE↓ \\
    \midrule
    fixed & fixed & 91.95  & 1.14  & 73.12  & 4.94  & 95.40  & 0.44  \\
    fixed & dynamic & 92.12 & 0.83  & 72.74  & \textbf{3.97} & 95.42  & 0.39  \\
    \rowcolor{gray!20} 
    dynamic & dynamic & \textbf{92.23}  & \textbf{0.76} & \textbf{73.21} & 4.21  & \textbf{95.47}  & \textbf{0.24} \\
    \bottomrule
    \end{tabular}
    }
  \caption{Ablation study on the impact of $\gamma$.}
  \label{tab:gamma}
\end{table}

\begin{figure}[htbp]
  \centering
  \resizebox{0.9\linewidth}{!}{
  \begin{subfigure}{0.32\linewidth}
    \includegraphics[width=1.0\linewidth]{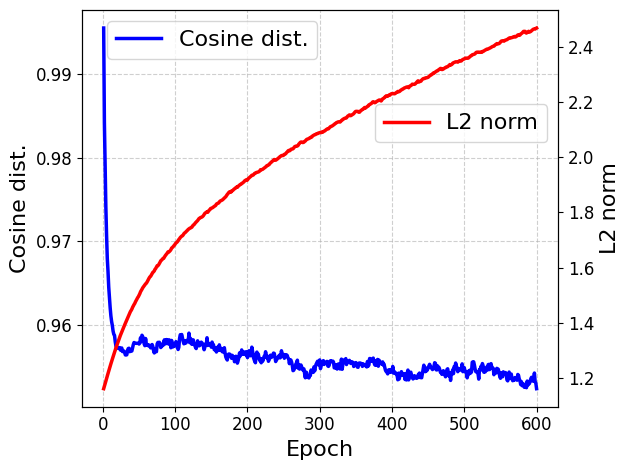}
    \caption{CIFAR-10}
  \end{subfigure}
  \hfill
  \begin{subfigure}{0.31\linewidth}
    \includegraphics[width=1.0\linewidth]{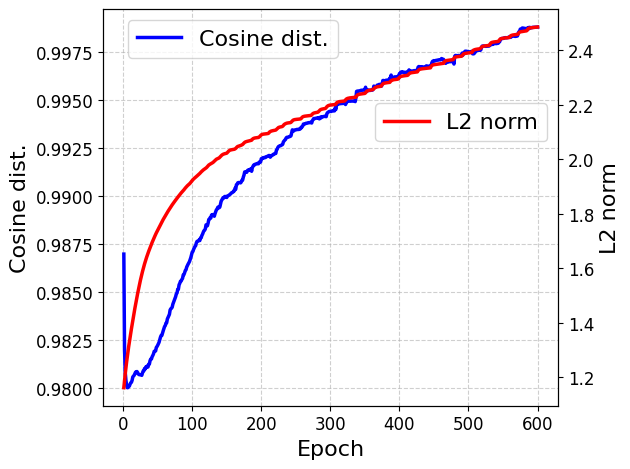}
    \caption{CIFAR-100}
  \end{subfigure}
  \hfill
  \begin{subfigure}{0.31\linewidth}
    \includegraphics[width=1.0\linewidth]{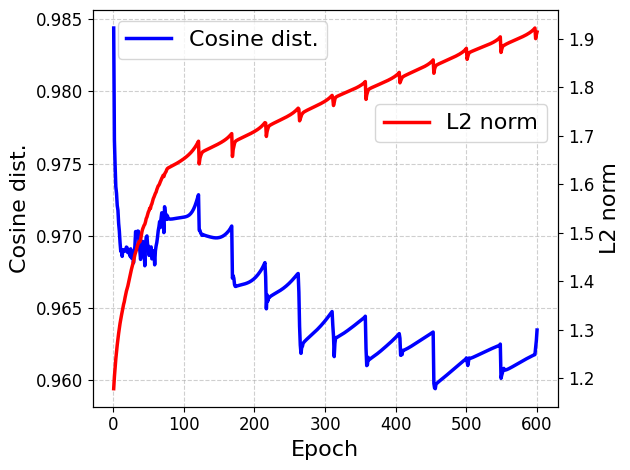}
    \caption{SVHN}
  \end{subfigure}
 }
  \caption{Dynamic changes of cosine and L2 distances between classifier prototypes during training.}
  \label{fig:distance}
\end{figure}

\paragraph{Evolution of prototype distances}
We analyze the evolution of prototype representations between two classifiers during training, as illustrated in \Cref{fig:distance}. For CIFAR-10 and SVHN datasets, the cosine distance between prototypes steadily decreased, indicating improved alignment of class representations, while the $L_2$ distance consistently increased, reflecting an expansion in prototype distributions. Conversely, the CIFAR-100 dataset exhibited a distinct trend: the cosine distance sharply declined in the early stages, followed by a gradual rise, likely due to the dataset's complex and dispersed class structures.
These observations highlight the significant impact of dataset-specific feature distributions on learning dynamics. For CIFAR-10 and SVHN, reduced cosine distance facilitated consistent high-level feature representations, enhancing class separability. Simultaneously, increased $L_2$ distance broadened prototype distributions, contributing to varied confidence outputs between classifiers. On CIFAR-100, the rapid initial alignment in cosine distance was followed by divergence, as seen in the increasing $L_2$ distance, signaling a more dispersed feature space.

By leveraging these dynamics, the growing $L_2$ distance, indicating the difference in output confidence between the two classifiers, serves as a foundation to balance learnable and ETF classifiers, refining calibration. Using a shared feature extractor and distinct learning strategies, the complementary interactions between classifiers improve model calibration, robustness, and accuracy.

\begin{figure}[htbp]
  \centering
    \includegraphics[width=1.0\linewidth]{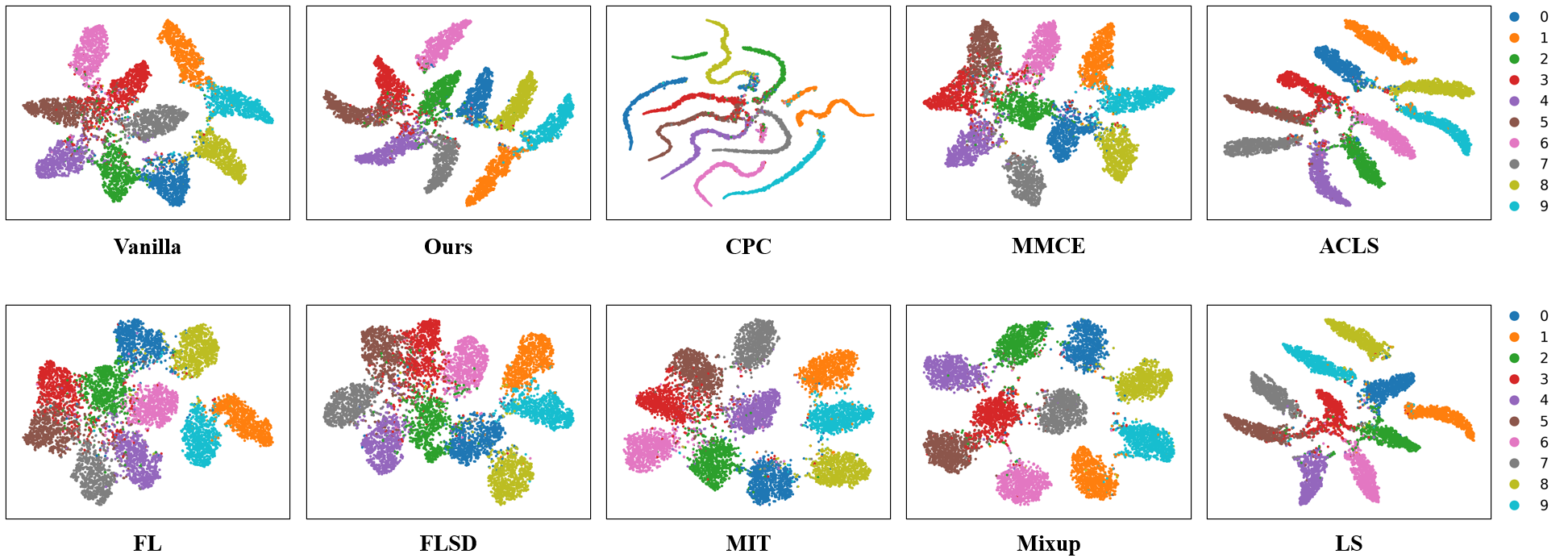}
    \caption{t-SNE visualizations of representations on CIFAR-10.}
  \label{fig:tsne2}
\end{figure}

\paragraph{Visualization results and analysis}
We present t-SNE visualizations of more training-time methods on CIFAR-10, as shown in \Cref{fig:tsne2}. We observe that our method yields well-clustered intra-class representations and more discriminative inter-class distances. This demonstrates that our approach effectively enhances representation learning and improves the balance between confidence and accuracy, which is crucial for addressing the overconfidence and underconfidence issues commonly encountered in model calibration.

\end{document}